\documentclass[journal]{IEEEtran}
\usepackage{color}
\usepackage[linesnumbered,boxed,ruled,commentsnumbered]{algorithm2e}
\usepackage{graphicx}
\usepackage{amsmath}
\usepackage[flushleft]{threeparttable}
\usepackage{footnote}
\usepackage{bm}
\usepackage{array}
\usepackage{makecell}
\usepackage{multirow}
\usepackage{booktabs}
\usepackage{bigstrut}
\usepackage{mathbbol}
\usepackage{xr}
\usepackage{subfigure}
\usepackage{adjustbox}

\newcommand{\tabincell}[2]{\begin{tabular}{@{}#1@{}}#2\end{tabular}}

\newcolumntype{M}[1]{>{\centering\arraybackslash}m{#1}}

\usepackage[switch]{lineno}
\begin{document}
	
%
	\title{DRL-FAS: A Novel Framework Based on Deep Reinforcement Learning for Face Anti-Spoofing}
%
%
%

	\author{Rizhao~Cai,
	        Haoliang~Li,
	        Shiqi~Wang,
	        Changsheng~Chen,
	        and~Alex~C.~Kot
	}

%


%

	\maketitle

\begin{abstract}
	Inspired by the philosophy employed by human beings to determine whether a presented face example is genuine or not, i.e., to glance at the example globally first and then carefully
	observe the local regions to gain more discriminative information, for the face anti-spoofing problem, we propose a novel framework based on the Convolutional Neural Network (CNN) and the Recurrent Neural Network (RNN). 
	In particular, we model the behavior of exploring face-spoofing-related information from image sub-patches by leveraging deep reinforcement learning. 
	We further introduce a recurrent mechanism to learn representations of local information sequentially from the explored sub-patches with an RNN. 
	Finally, for the classification purpose, we fuse the local information with the global one, which can be learned from the original input image through a CNN. 
	Moreover, we conduct extensive experiments, including ablation study and visualization analysis, to evaluate our proposed framework on various public databases. 
	The experiment results show that our method can generally achieve state-of-the-art performance among all scenarios, demonstrating its effectiveness.
\end{abstract}


\IEEEpeerreviewmaketitle

\section{Introduction}\label{sec-introduction}
\IEEEPARstart{F}{ace} recognition techniques have been increasingly deployed in everyday scenarios for authentication purposes, such as mobile devices unlocking and door control access.
Compared with other biometric information, using faces for authentication is more user-friendly as face verification is non-intrusive, and the face images can be feasibly captured with mobile phone cameras.
However, it has been widely recognized that state-of-the-art face recognition systems are still vulnerable to spoofing attacks. Attackers can easily hack a face recognition system by presenting a spoofing face of a client to the system's camera, where a spoofing face could be a face mask and a face image shown by a printed photo or by a digital display.
Therefore, reliable Face Anti-Spoofing (FAS) techniques are highly desired and essential for developing secure face recognition systems.

\begin{figure}[t]
	\centering
	\includegraphics[width=1.0\linewidth]{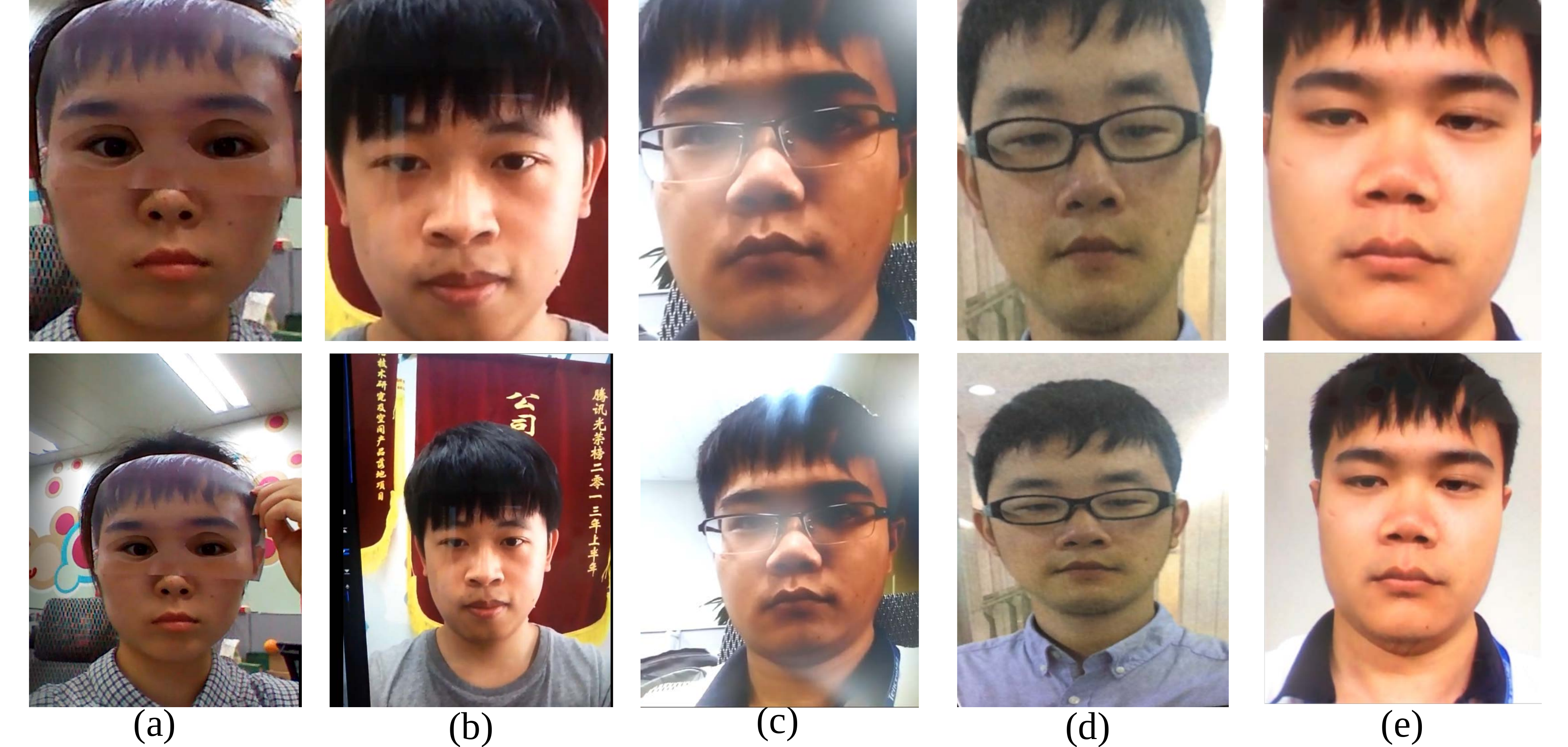}
	\caption{Presentation attack examples in different scenarios. The examples are from the ROSE-YOUTU Face Liveness Detection Database \cite{FAS-UnsupervisedDA-TIFS-2018}. The spoofing face examples are shown at the bottom of each column, and the cropped faces are shown at the top. (a) an example of paper mask which is not close enough to the acquisition camera such that the paper boundary can be spotted. (b) an example of video attack which is not close enough to the acquisition camera such that the bezel of the presentation medium (in the left) can be seen. (c) an example of video attack which is acquired with a camera close enough. Although no bezel is visible, obvious spoofing clues (e.g. reflections) appear around the face. (d) an example of print attack. (e) an example of replay attack. The examples in (d)\&(e) are launched carefully such that no paper boundary, bezel and obvious spoofing clues can be seen.}\label{fig-examples}
	\hspace{-5cm}
\end{figure}
  
\par The past few years have witnessed much progress in the FAS problem.
Traditionally, in either the Spatial or Fourier space, various techniques have been proposed to extract handcrafted features with image descriptors as representations \cite{FAS-CoALBP-AIVT-2012P,DB-IDIAP-RA,FAS-CoALBP-AIVT-2012P,FAS-ColorTexture-TIFS-2016,DoG-ECCV-2010,FAS-LPQ-TIFS-2015}. 
These features are usually used to train a Support Vector Machine (SVM) to classify genuine or spoofing examples. 
However, these features are insufficiently discriminative because those descriptors (e.g., Local Binary Pattern) are not originally designed for the FAS problem. 
\par Recently, deep-learning-based methods, which aim to learn discriminative representations in an end-to-end manner, have shown evidence to be more effective in countermeasures against spoofing attacks than the traditional methods.
Yang \textit{et al.} \cite{FAS-CNN-ComputerScience-2014} firstly introduce the Convolutional Neural Network (CNN) for the FAS task.
They train an AlexNet-based model \cite{CV-AlexNet-CVPR-2012}, extract features from the model's last layer and learn an SVM with binary labels (``genuine'' or ``spoofing'') for classification. 
Besides using binary labels, Liu \textit{et al.} \cite{FAS-Auxiliary-CVPR-2018} seek for the auxiliary supervision signals.
They use auxiliary techniques to extract pseudo depth maps and remote PhotoPlethysmoGraphy (rPPG) signals from the RGB images for supervision to boost the training.
It is also reported that the Recurrent Neural Network (RNN) can be used to utilize temporal information from sequential frames for face anti-spoofing \cite{FAS-LSTMCNN-ICASSP-2018, FAS-CVPR2019-STASN}.
However, one limitation of the aforementioned techniques is that the learned feature representations may overfit to the properties of a particular database. For example, depth information can benefit face anti-spoofing when the suspicious input is in 2D format (e.g., printed photo, screen display), but it is likely to fail to counter mask attacks, which are with 3D information (e.g., Fig.~\ref{fig-examples}(a)). 
To learn more spoofing-discriminative representations and alleviate the overfitting effect, we propose a novel two-branch framework based on CNN and RNN. 

    \begin{figure}[t]
    	\centering
    	\includegraphics[width=1.0\linewidth]{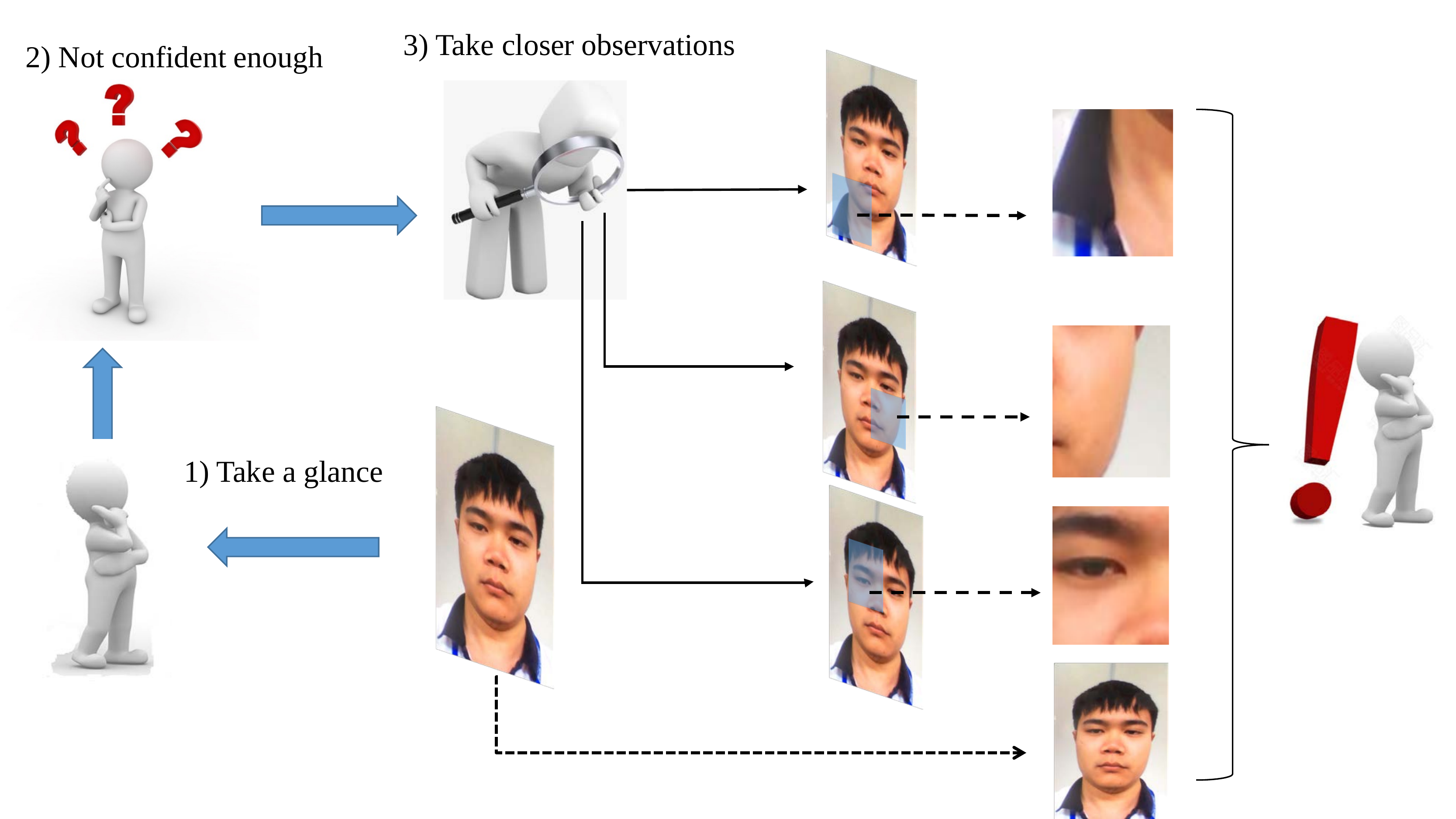}
    	\caption{Illustration of how a testee  assesses the liveness of a face example without any apparent distortion. 1) Firstly, the testee will take a glance at the given face example. 2) The testee may not tell that it is a spoofing face with certainty because it looks similar to a genuine face. 3) To confirm the assessment, the testee will take closer observations to carefully search subtle spoofing clues. After several steps of observations, the testee can provide a more accurate assessment by increasingly discovering spoofing clues. Therefore, having observed the example globally and locally, the testee can make a more reliable assessment than a glance without any closer observation.}\label{fig-human-behavior}
    	\hspace{-5cm}
    \end{figure}

    \subsection{Motivation}\label{sec-introduction-motivation}
    \par The motivations behind this work are inspired by 1) the observation that spoofing clues can appear in various ways and 2) how human beings can act to predict whether a presented face example is genuine or spoofing. 
    Regarding the first motivation, we show motivating examples in Fig.~\ref{fig-examples}, which indicates that spoofing clues can be diverse.
    In occasional cases, spoofing clues are visually salient, such as the boundaries of a printed photo, the bezel of a digital display \cite{patel2016secure}, and reflections, which are respectively shown in Fig.~\ref{fig-examples}(a), Fig.~\ref{fig-examples}(b) and Fig.~\ref{fig-examples}(c).
    These clues can be easily spotted and used by human beings to assess the examples as ``spoofing'' even without further careful observation.
    However, in most cases, human beings may give prediction with less certainty as the aforementioned clues may be inconspicuous if an attacker carefully launches the spoofing attack.
    For instance, no paper boundary, bezel, or reflection appears in Fig.~\ref{fig-examples}(d) and Fig.~\ref{fig-examples}(e).
    Moreover, the visual quality of Fig.~\ref{fig-examples}(d) and Fig.~\ref{fig-examples}(e) is better than that of Fig.~\ref{fig-examples}(a), Fig.~\ref{fig-examples}(b) and Fig.~\ref{fig-examples}(c).
    In other words, Fig.~\ref{fig-examples}(d) and Fig.~\ref{fig-examples}(e) look more similar to genuine faces, and thus human beings may not tell the difference with only a glance.
    Thus, to mine more information to confirm their assessment, human beings would carefully delve into local sub-patches to explore fine-scale and subtle spoofing clues. Fig.~\ref{fig-human-behavior} portrays such behavior, which reveals our second motivation.

    \par Under these two motivations, we propose a two-branch framework, DRL-FAS, that jointly exploits global and local features based on CNN, RNN, and deep reinforcement learning (DRL), for the face anti-spoofing (FAS) problem.
    Fig.~\ref{fig-framework} elaborates on this framework, which corresponds to Fig.~\ref{fig-human-behavior}. 
    Firstly, we treat human beings' glance at an example as the procedure of extracting global features. As such, we train a CNN to learn global information through the entire frames from video data.
    Then, we treat the following closer observations at suspicious sub-patches as the procedure of extracting local features.
    To model such observation behavior, we leverage reinforcement learning to learn a policy model that predicts locations of suspicious sub-patches and learn local information there with RNN.
    Finally, since human beings can benefit from both global and local information for better prediction, the extracted global and local features are fused for classification.

    \par The contributions of this work are three-fold:\\
    $\bullet$ We propose a novel framework based on CNN and RNN for the FAS problem. Our framework aims to extract and fuse the global and local features.
    While many of the previous works used RNN to leverage temporal information from video frames, we take advantage of RNN to memory information from all ``observations" from sub-patches to reinforce extracted local features gradually.\\
    $\bullet$ To explore spoofing-specific local information, we leverage the advantage of reinforcement learning to discover suspicious areas where discriminative local features can be extracted.
    To the best of our knowledge, in the field of FAS, this is the first attempt to introduce reinforcement learning for the optimization.\\
    $\bullet$ We conduct extensive experiments using six benchmark databases to evaluate our method.
    As shown in Section \ref{sec-exp-results}, our method can perform better than the schemes that either use global or local features. Moreover, our proposed method can generally achieve state-of-the-art performance compared with other methods.

\section{Related Works} \label{Sec_RelatedWorks}

 \subsection{Traditional Face Anti-Spoofing} \label{S2-1-1}
Most of the traditional FAS techniques focus on designing handcrafted features and learning classifiers with learning methods such as SVM.
Texture-based methods are based on the assumption that there are differences in texture between genuine faces and spoofing faces due to the inherent nature of different materials. 
In the Fourier spectrum, Tan \textit{et al.} \cite{DoG-ECCV-2010} propose to use Difference-of-Gaussian (DoG) features to describe the frequency disturbance caused by the recapture.
Besides, Gragnaniello \textit{et al.} \cite{FAS-LPQ-TIFS-2015} propose to use Local Phase Quantization (LPQ) to analyze texture distortion through the phase of images.
Also, texture descriptors, such as Local Binary Pattern (LBP), Scale-Invariant Feature Transform (SIFT), are used in the spatial domain to extract features to represent such disparities \cite{FAS-MicroTexture-IJCB-2011,FAS-CoALBP-AIVT-2012P,CDD-ICB-2013,  FAS-FisherVector-TIFS-2017, FAS-ColorTexture-TIFS-2016}.
In addition, due to the distortion caused during the recapture process, spoofing examples usually have lower visual quality compared with genuine ones. Motivated by this observation, the FAS community also has proposed to detect spoofing attack examples by assessing the input image quality \cite{IQA-ICPR-2014,FAS-IDA-TIFS-2015, LI-QUALITY}.
\textcolor{black}{Apart from analyzing a single image, methods based on sequential video frames are also proposed to utilize the information from the temporal space, such as eye blinking, lip moves~\cite{Motion-IJCB-2011,MotionLBP-ICB-2013,LBP-TOP-EJIVP-2014}, and motion blurring effect \cite{TIFS-2019-MotionBlure}.} Although such methods can be effective against photo attacks, they cannot counter video replay attacks where such movement information can exist in a given video.

\subsection{Deep-Learning-Based Face Anti-Spoofing}
\par Recently, deep learning has dominated the computer vision community, so as the FAS field.
Yang \textit{et al.} \cite{FAS-CNN-ComputerScience-2014} are the first to apply CNN to the FAS problem.
The authors train a deep model based on AlexNet \cite{CV-AlexNet-CVPR-2012} architecture and extract features from the model's last layer to train an SVM classifier.
However, this is just a straightforward application of AlexNet, and the improvement is limited compared with the handcrafted features.
After that, more CNN-based methods are proposed \cite{BottleCNN-JVCIR-2016,FAS-MSR-TIFS-2019, FAS-MultichannelCNN-TIFS-2020}.
Moreover, RNN (e.g., Long Short-Term Memory networks \cite{ML-LSTM-NC-1997}, Gated Recurrent Unit \cite{ML-GRU-ArXiv-2014}) can also be used for the FAS problem by leveraging temporal information from sequential video frames \cite{FAS-LSTMCNN-ACPR-2015, FAS-LSTMCNN-ICASSP-2018}. 
By far, the aforementioned methods merely use binary labels (``genuine" and ``spoofing") for training. Other than that, the methods in \cite{FAS-Depth&Patch-IJCB-2018,FAS-Auxiliary-CVPR-2018, FAS-DeSpoofing-ECCV-2018} utilize extra information for auxiliary supervision, such as depth. 
For example, Atoum \textit{et al.} \cite{FAS-Depth&Patch-IJCB-2018} introduce auxiliary depth information for spoofing detection. They hold the idea that 2D spoofing examples of printed papers and screens are flat and thus lack 3D information. As such, they train a CNN-based depth estimator, by which the output toward genuine and spoofing faces are 3D depth maps and flat maps, respectively.
Although the performance is shown to be improved with depth as auxiliary supervision, such depth-based methods may not work efficiently when a paper mask attack with depth information is launched.
 
\begin{figure}[t]
	
	\includegraphics[height=0.5\linewidth]{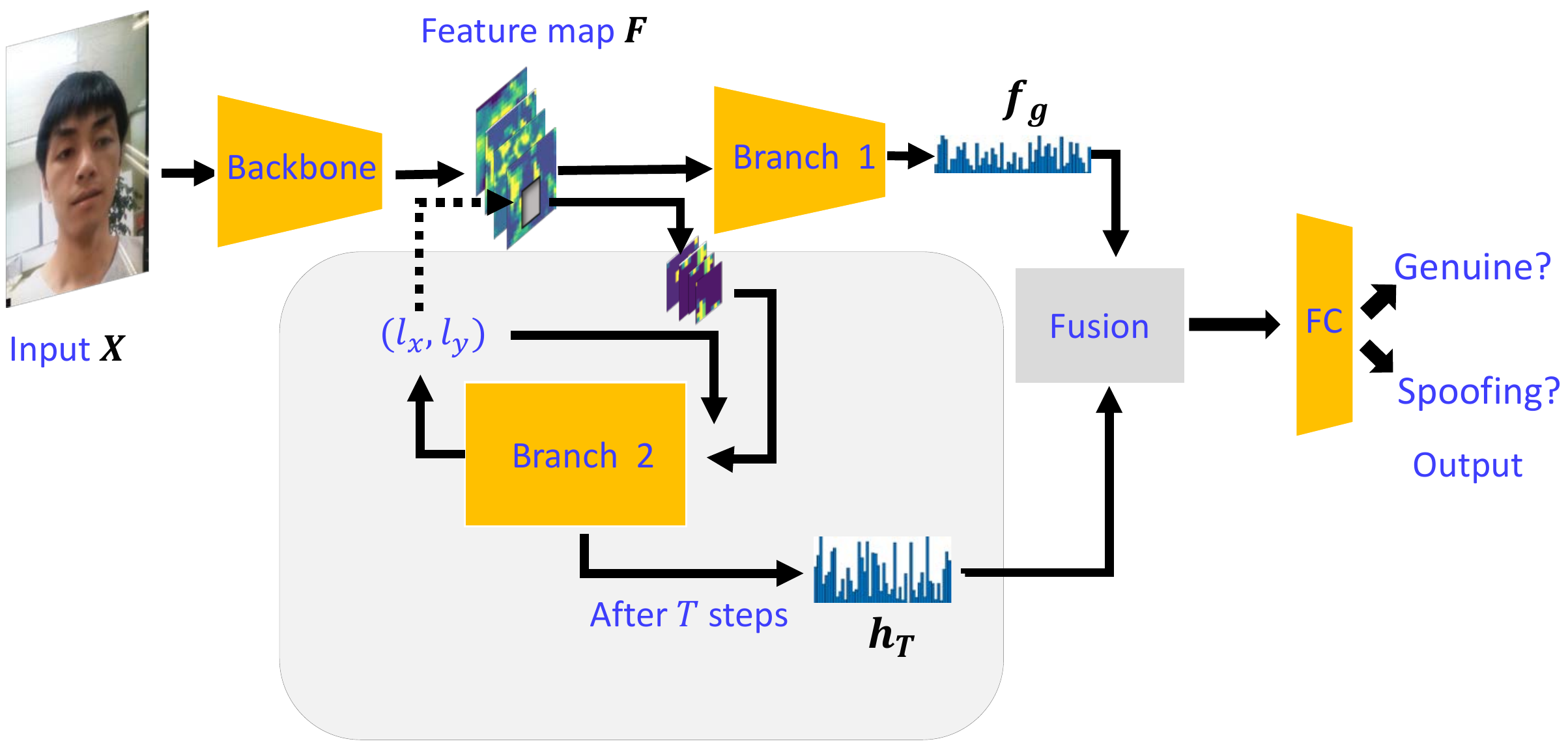}
	\caption{Overview of the proposed framework. The backbone and Branch 1 networks are based on CNN, and Branch 2 is based on RNN. The working pipeline is as follows: 1) The backbone network processes the input \bm{$X$} into the feature map \bm{$F$}. 2) Subsequently, Branch 1 extracts global features from the entire \bm{$F$}. Meanwhile, Branch 2 extracts local features from sub-patches of \bm{$F$} recursively. 3) Finally, the extracted global and local features are fused for the classification in the last fully-connected (FC) layer.}\label{fig-framework}
\end{figure}

\subsection{Cross-Domain Face Anti-Spoofing}
The variety of capturing settings, such as different cameras, environment illuminations, presentation mediums, etc., can lead to the domain shift problem \cite{FAS-UnsupervisedDA-TIFS-2018}. To be specific,
a model trained with data collected under one condition setting may not be able to generalize to other settings. 
This problem deters models from being deployed in practical scenarios.
Aimed at making models more generalized and overcoming the domain shift problem, transfer-learning-based algorithms regarding either domain adaptation/generalization or zero/few-shot learning are also proposed \cite{FAS-UnsupervisedDA-TIFS-2018, FAS-3DCNN-TIFS-2018, FAS-CVPR-SHAO-M, FAS-ZeroShot-CVPR2019, LI-DISTILLATION, LIZHI}. However, it is still an open problem regarding how to design a sophisticated transfer learning algorithm for FAS by considering all possible capturing settings.

\section{Methodology}\label{sec-method}
\begin{figure}[t]
	\centering
	\includegraphics[height=0.5\linewidth]{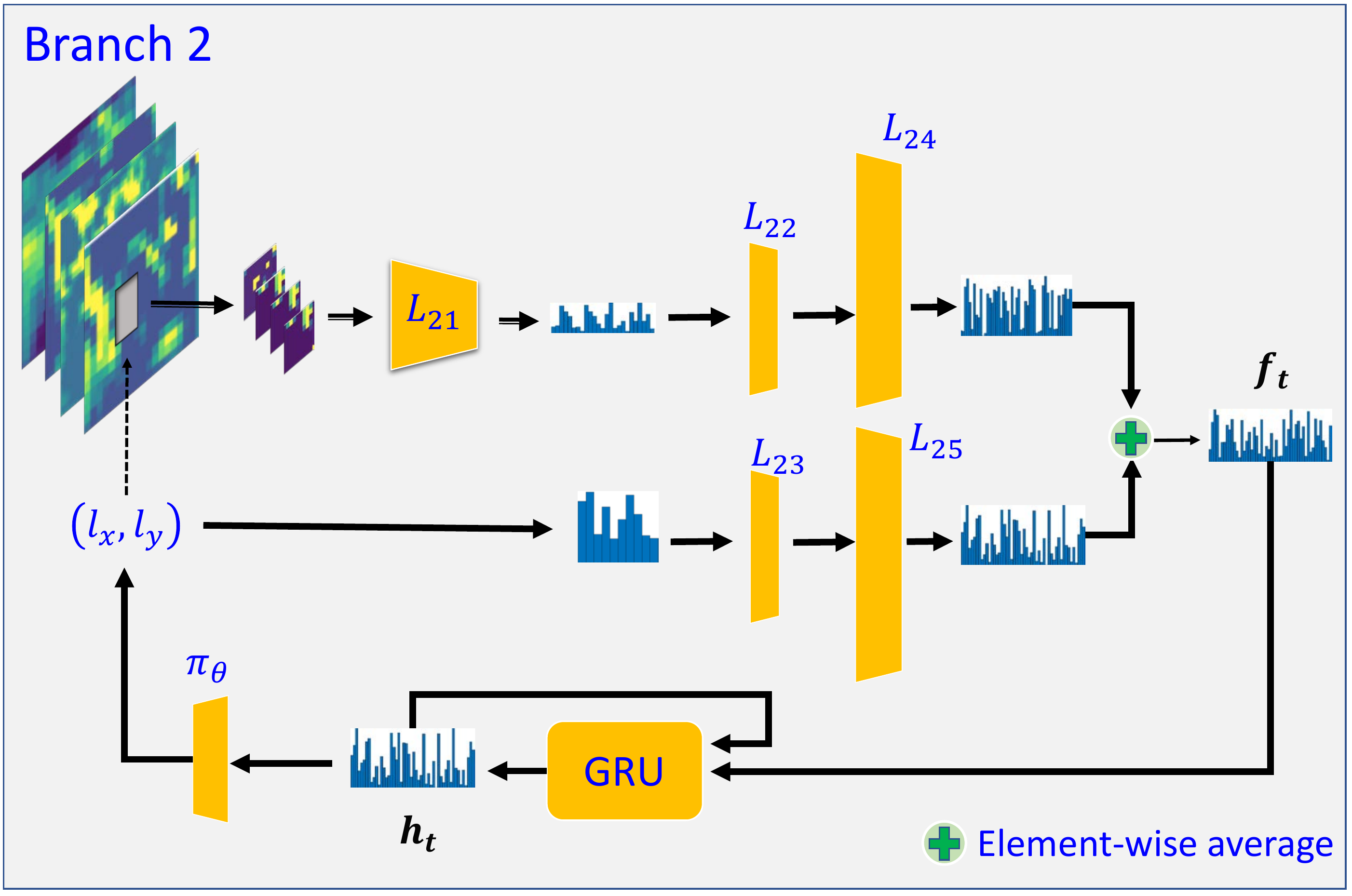}
	\caption{Illustration of Branch 2. It consists of a Gated Recurrent Unit (GRU) and several linear layers ($\pi_{\bm{\theta}}$, $L_{21}$, $L_{22}$, etc.). It takes $T$ steps for local features to be extracted. For each step, the reinforcement learning agent predicts a location $(l_{x_t}, l_{y_t})$ via its policy model ($\pi_{\bm{\theta}}$). Subsequently, at $(l_{x_t}, l_{y_t})$, a sub-patch from \bm{$F$} will be cropped as \bm{$F_{(l_{x_t},l_{y_t})}$}, and \bm{$F_{(l_{x_t},l_{y_t})}$} will be further processed by the GRU. After $T$ steps, the output hidden state of the GRU ($\bm{h_T}$) serves as the extracted local features to be fused with the global features for the classification purpose.}\label{fig-branch2}
\end{figure}

In this work, we propose a two-branch framework based on CNN, RNN inspired by how human beings can act to observe and explore spoofing clues. 
The overview of the framework is shown in Fig.~\ref{fig-framework}. The backbone network firstly embeds the RGB image $\bm{X}$ into a feature map \bm{$F$}. 
Then, \bm{$F$} is forwarded to the subsequent Branch 1 and Branch 2 for the extraction of global and local discriminative features, respectively.
Finally, these features are fused for the classification purpose in the final fully-connected (FC) layer. The details of Branch 2 are shown in Fig.~\ref{fig-branch2}, where $L_{21}$ is a convolutional layer, and $L_{22}$, $L_{23}$, $L_{24}$ and $L_{25}$ are fully-connected layers. 
In Section \ref{sec-method-global}, we describe how the backbone and Branch 1 work cooperatively to extract global features. 
Subsequently, in Section \ref{sec-method-local}, we illustrate the extraction of local features from suspicious sub-patches sequentially. Afterward, the reinforcement learning leveraged to predict the locations of those sub-patches will be elaborated in Section \ref{sec-method-rl}. 
Finally, in Section \ref{sec-method-training}, we present the optimization process.

\subsection{Global Feature Extraction} \label{sec-method-global}
The global feature extraction aims to exploit global discriminative information (e.g., paper boundaries, bezels, salient reflection patterns).
In our framework, global features are extracted in Branch 1, a sequence of convolutional blocks. 
Branch 1 processes \bm{$F$} from the backbone to extract global features $\bm{f_g}$, into which all elements from \bm{$F$} are encoded. 

\par For implementation, we construct the backbone and Branch 1 based on the ResNet18 \cite{CV-He-2016-ResNet} architecture such that we can fairly compare our method with the recent ResNet-based methods (e.g., \cite{FAS-MSR-TIFS-2019}).
In particular, we adopt the first convolutional layer and four subsequent residual blocks of ResNet18 as our backbone network. 
The remaining convolutional residual blocks and the Global Average Pooling (GAP) layer \cite{ML-GAP-ICLR-2014} constitute Branch 1. 
The details of these modules are provided in Table \ref{tab-parameters}.

\subsection{Local Feature Extraction} \label{sec-method-local}
\par The local features are expected to exploit discriminative information from a local patch.
The local feature extraction consists of $T$ steps in total. We first elaborate on the procedure of local feature extraction at a certain step $t$. At the beginning of step $t$, a location $(l_{x_t}, l_{y_t})$ is first produced, where $l_{x_t}$ and $l_{y_t}$ represent the horizontal and vertical coordinates respectively.
The learning of the location prediction is via reinforcement learning, which will be introduced in the next sub-section. Then, from \bm{$F$}, a square sub-patch \bm{$F_{(l_{x_t},l_{y_t})}$} with the patch size $p$ centering at $(l_{x_t}, l_{y_t})$ is cropped.
Next, \bm{$F_{(l_{x_t},l_{y_t})}$} together with the location information $(l_{x_t}, l_{y_t})$ will be encoded to an intermediate feature \bm{$f_t$}. As such, \bm{$f_t$} contains information from the observation at step $t$. 

\par While previous works usually utilize RNN to leverage temporal information from sequential video frames, we particularly employ a Gated Recurrent Unit (GRU \cite{ML-GRU-ArXiv-2014}) to learn local features from $\bm{f_1}, \bm{f_2}, \ldots \bm{f_t}$ in a sequential and recursive manner. 
The reason is that the hidden state $\bm{h_t}$ of the GRU can be learned from $\bm{h_{t-1}}$ and $\bm{f_{t-1}}$:
    \begin{equation}
        \begin{aligned}
            \bm{z_t} =& sigmod(\bm{W_z} \bm{f_t} + \bm{U_z} \bm{h_{t-1}} + \bm{b_z})\\
            \bm{q_t} =& sigmod(\bm{W_q} \bm{f_t} + \bm{U_q} \bm{h_{t-1}} + \bm{b_q}) \\
             \hat{\bm{h_t}} =& \tanh{(\bm{W_h} \bm{f_t} + \bm{U_h}(\bm{q_t} \odot \bm{h_{t-1}}) + \bm{b_h}})\\
            \bm{h_t} =& \bm{z_t} \odot \bm{h_t} + (1-\bm{z_t}) \odot \hat{\bm{h_t}},
        \end{aligned}
    \end{equation}
where $\odot$ is the Hadamard product operation, $\bm{W_{\{z,q,h\}}}$, $\bm{U_{\{z,q,h\}}}$, and $\bm{b_{\{z,q,h\}}}$ are parameters of the GRU. 
Furthermore, when analyzing a presented example, human beings' knowledge with respect to the example grows as information is gradually gained at each step. 
In other words, based on previous observations ($\bm{f_1}, \bm{f_2}, \ldots \bm{f_t}$), their assessment toward a suspicious example can get reinforced after a new observation ($\bm{f_{t+1}}$).
Therefore, we specifically employ the GRU to memorize the observed information and learn local features.   
After $T$ steps, $\bm{h_T}$ is treated as the final extracted local feature because it has perceived the local information during the $T$ steps of observations. Finally, the proposed framework jointly exploits global and local features by fusing $\bm{h_T}$ and $\bm{f_g}$ for the classification purpose.

\subsection{Reinforcement Learning for Face Anti-Spoofing} \label{sec-method-rl}
    To explore spoofing-discriminative local information, we leverage reinforcement learning to train an agent that can help predict locations of sub-patches where spoofing clues may appear. In this context, a reinforcement learning agent is an abstract subject that explores clues in a certain environment and predicts locations. 
    \par \textbf{Environment: } In our framework, we treat \bm{$F$} from the backbone as the environment where our agent predicts locations and gets feedbacks to update its policy.
    This is because the backbone can distill shallow spoofing-related features from raw RGB pixels into \bm{$F$}. Thus, \bm{$F$} can especially provide spoofing-related information for the agent to predict appropriate locations to extract effective local features.
    \par Although an input RGB image (\bm{$X$}) can also be set as the environment, we experimentally find that using the backbone to extract \bm{$F$} and setting it as the environment can provide better results. 
    We conjecture that raw pixels of an RGB image may contain interference such that the agent could get overwhelmed in a complex environment. 
    On the other hand, the backbone could filter out unrelated information and distill spoofing-related information from raw pixels, and thus provide a specific environment \bm{$F$} for the agent to explore spoofing clues with less disturbance.
    The experimental results in \ref{sec-exp-ablationstudy} show the superiority of using the backbone and setting \bm{$F$} as the environment.

    \par \textbf{State: } At step $t$, according to a certain policy, the agent predicts the location of a sub-patch based on its state $s_t$.
    As $\bm{h_t}$ carries history action information, we define the agent's state as $\bm{h_t}$: $s_t=\bm{h_t}$.

    \par \textbf{Action and Policy: }
    In the framework, the agent learns to predict the location of a sub-patch to explore spoofing information. Hence, our agent's action $a_t$ is to predict the location: $a_t=(l_{x_t},l_{y_t})$. 
    Then, the sub-patch at $(l_{x_t},l_{y_t})$ will be cropped for the extraction of local features.
    \par For effective location prediction, an optimal policy $\pi$ should be provided to guide our agent to predict a location according to its current state: $a_{t}=\pi(s_t)$. 
    Following the policy gradient theory \cite{ML-2000-PolicyGradient}, we parameterize $\pi$ as $\pi_{\bm{\theta}}$ by using a differentiable linear layer, where $\bm{\theta}$ denotes its parameters. In this way, we can optimize
    $\bm{\theta}$ with the standard backward propagation based on reward signals, which will be illustrated later (Section \ref{sec-method-training}).

    \par \textbf{Reward: } After predicting the locations, the agent should get reward signals to evaluate how discriminative the information that the sub-patches contain for the classification. The more effective the predictions, the higher the rewards. 
    Since the classification is conducted at the final step, we define a delayed reward as
    \begin{equation}
        r_t=\left\{
            \begin{array}{lr}
            0, & \text{if } t<T\\
            {\rm log} P(y_{gt}|\bm{X}), &\text{if } t=T, \\
            \end{array}
        \right.
    \end{equation} 
    where $r_t$ is the reward signal at step $t$, $T$ is the total number of observation steps, $y_{gt}$ is the ground-truth label of \bm{$X$}, and $P$ is the predicted probability distribution over the binary labels.
    The agent will be trained to gain the cumulative reward 
    \begin{equation}\label{eq-reward}
        R= \Sigma_{t=1}^{T} r_t = {\rm log} P(y_{gt}|\bm{X})
    \end{equation}
    as high as possible.

   \subsection{Training and Optimization} \label{sec-method-training}

          \subsubsection{Two-stage training scheme}
            \par When optimizing our framework, although end-to-end training is achievable, it may not provide satisfactory performance.
            As mentioned, the output feature map \bm{$F$} from the backbone can be seen as an environment where our agent acts to learn its policy.
            If the backbone is involved in training, the environment will be unstable.
            Assume that the training is in epoch $i$, $\pi_{\bm{\theta}}$ is optimized according to the ``environment'' $\bm{F}_{i}$.
            However, if the backbone is also involved in optimization, the ``environment" will change in the next epoch, meaning that $\bm{F}_{i+1} \neq \bm{F}_{i}$.
            Therefore, the agent may not act properly to the new ``environment" $\bm{F}_{i+1}$ with the policy learned from the $\bm{F}_{i}$.
            \par To tackle this problem, we propose to use a two-stage training scheme.
            At the first stage, we pretrain a ResNet18 model with the training data.
            Then, the parameters of the corresponding modules will be loaded to the backbone from the pretrained model.
            Subsequently, the parameters of the backbone will be frozen and not involved in the second-stage optimization such that $\bm{F}_{i+1}=\bm{F}_{i}$.
            As such, fixing the parameters of the backbone is to fix \bm{$F$}, which can help keep a stable ``environment" and extract more effective local features.
           The experimental results in Section \ref{sec-exp-results} show the superiority of our two-stage training.
  		\subsubsection{Joint optimization}
            At the second stage, we optimize the parameters of Branch 1 and 2 jointly.
            The parameters of Branch 1 and Branch 2 except $\pi_{\bm{\theta}}$ are optimized by the standard cross-entropy loss with binary labels for supervision.
            \par As for $\pi_{\bm{\theta}}$, it is optimized with reinforcement learning. The optimization of $\pi_{\bm{\theta}}$ can be formulated as the maximizing of the following objective function:
            \begin{equation}
                J(\bm{\theta}) = \mathbb{E}_{\rho(s_{1:T}; \bm{\theta})}[R],
            \end{equation}
        where $\rho(s_{1:t}; \bm{\theta})$ is the distribution over action sequences, which depends on $\pi_{\bm{\theta}}$.
  		
        \par According to the policy gradient theory \cite{ML-2000-PolicyGradient}, we adopt a differentiable linear layer to approximate the policy function.
        Hence, the maximization of $J(\bm{\theta})$ can be via the calculation of the gradient of $J(\bm{\theta})$ and the application of the gradient ascend.
        To this end, we leverage the REINFORCE rule \cite{ML-REINFORCE} to approximate the gradient of $J(\bm{\theta})$:
  		\begin{equation}
  		\begin{aligned}
	  		\nabla_{\bm{\theta}} J(\bm{\theta})&=\sum_{t=1}^{T}\mathbb{E}_{\rho(s_{1:t};\bm{\theta})}[\nabla_{\bm{\theta}} \log\pi_{\bm{\theta}}(a_t|s_{1:t};\bm{\theta})R] \\
	  		&\approx \sum_{t=1}^{T}\nabla_{\bm{\theta}} \log\pi_{\bm{\theta}}(a_t|s_{1:t};\bm{\theta})R.
  		\end{aligned}
  		\end{equation}
        As the gradient of $J(\bm{\theta})$ can be simply computed by $ \nabla_{\bm{\theta}} \log\pi_{\bm{\theta}}(a_t|s_{1:t};\bm{\theta}$).
        Thus, $\pi_{\bm{\theta}}$ can be optimized by the standard backward propagation with this approximated gradient.
  		
\section{Experiments} \label{Sec-exp}
This section describes how we conduct experiments to evaluate our method.
To begin with, we introduce six benchmark databases employed in the experiments.
After that, we illustrate the implementation details. 
Finally, we present and discuss the experimental results.

\subsection{Databases}\label{sec-exp-db}
We utilize six publicly available face presentation attack databases in our experiments, including CASIA Face Anti-Spoofing Database \cite{DB-CASIAFASD}, IDIAP REPLAY-ATTACK \cite{DB-IDIAP-RA}, MSU Mobile Face Spoofing Database \cite{FAS-IDA-TIFS-2015}, OULU-NPU database \cite{OULU_NPU_2017}, the Spoofing in the Wild (SiW) database \cite{FAS-Auxiliary-CVPR-2018} and the ROSE-YOUTU database \cite{FAS-UnsupervisedDA-TIFS-2018}.

\subsubsection{CASIA FASD}
The CASIA Face Anti-Spoofing Database (CASIA for short) has 20 and 30 subjects in its training and testing set respectively.
For each subject, there are 12 videos, among which 3 genuine face videos are recorded from the genuine faces and 9 spoofing face videos from photos and screens. As such, the CASIA database has 600 videos in total. 

\subsubsection{IDIAP REPLAY-ATTACK}
The IDIAP REPLAY-ATTACK database \cite{DB-IDIAP-RA} (IDIAP for short) is constituted of 1,200 videos in total, with 360, 360 and 480 videos in the training set, development set, and testing set, respectively.
In this database, there are two illumination conditions: 1) a controlled condition where the background is uniform and the source of lighting is a fluorescent lamp; 2) an adverse condition where the background is non-uniform and the source of lighting is daylight.
The attack videos of each subject involve the 1) Print Attack: High-resolution face pictures printed on a paper. 2) Replay Attack: High-resolution pictures or videos were displayed on the screen of an iPhone 3GS and an iPad.
To collect such data, the webcam of a MacBook, an iPhone 3GS and a Canon PowerShot camera are used.

\subsubsection{MSU MFSD}
The MSU Mobile Face Spoofing Database (MSU for short) \cite{FAS-IDA-TIFS-2015} consists of 280 video clips of photo and video attack from 35 subjects.
Two types of cameras are used to collect videos: the built-in camera in MacBook Air 13” ($640\times480$) and the front-facing camera in the Google Nexus 5 Android phone ($720\times480$).
However, all the videos are only collected in normal indoor lighting environments.

\subsubsection{OULU-NPU}
The OULU-NPU database, similar to the IDIAP database, is divided into the training, development, and testing set with 20, 15, and 20 subjects, respectively. 
Overall, it contains 4950 face videos collected under three different environment conditions (e.g., different illumination and background conditions), with the frontal cameras of six mobile phones.
As for attack mediums, two printers and two display devices were used to produce print attack and video attack examples.
Furthermore, the OULU-NPU database provides four protocols for evaluation. Among them, Protocol 1, 2, and 3 aim to evaluate a model's generalization capability to unseen environment conditions, unseen attack mediums, and unseen camera modules, respectively. Protocol 4 simultaneously considers the unseen environment conditions, attack mediums, and camera modules.

\begin{table}[tbp]
	\centering
	\caption{The parameters of each module in the framework. ``2D conv'' denotes the a sequence of a 2D convolutional layer, a 2D batch normalization layer and a ReLU layer. ``Linear'' denotes a sequence of a fully-connected layer and a ReLU layer. ``GAP'' denotes the Global Average Pooling layer. The input and output sizes of each layer are shown. {$p$} is the patch size for the cropping.
	}
	\begin{tabular}{|c|c|c|c|}
		\hline
		Module & Layer & Type  & Size \bigstrut\\ \hline
		\multirow{4}[8]{*}{Backbone}          & 1     & 2D Conv  &  \tabincell{c}{In: $3\times256\times256$\\Out: $64\times256\times256$}  \bigstrut\\
		\cline{2-4}          & 2     & MaxPooling  & \tabincell{c}{In: $64\times256\times256$\\Out: $64\times128\times128$} \bigstrut\\
		\cline{2-4}          & 3     & Residual Block $\times 2$ & \tabincell{c}{In: $64\times128\times128$\\Out: $64\times64\times64$},  \bigstrut\\
		\cline{2-4}          & 4     &  Residual Block $\times 2$ & \tabincell{c}{In: $64\times64\times64$\\Out: $128\times32\times32$} \bigstrut\\
		\hline
		\multirow{3}[6]{*}{Branch 1} & 1     & Residual Block $\times 2$ & \tabincell{c}{In: $128\times32\times32$\\Out: $256\times32\times32$} \bigstrut\\
		\cline{2-4}          & 2     & Residual Block $\times 2$ & \tabincell{c}{In: $256\times32\times32$\\Out: $512\times32\times32$} \bigstrut\\
		\cline{2-4}          & 3     & GAP   &  \tabincell{c}{In: $512\times32\times32$\\Out: $512\times1\times1$} \bigstrut\\
		\hline
		\multirow{7}[14]{*}{Branch 2} & $L_{21}$   & 2D Conv + GAP & \tabincell{c}{In: $128\times p\times p$\\Out: $256\times1\times1$} \bigstrut\\
		\cline{2-4}          & $L_{22}$   & Linear & \tabincell{c}{In: $256$\\Out: $512$} \bigstrut\\
		\cline{2-4}          & $L_{23}$   & Linear & \tabincell{c}{In: $2$\\Out: $512$} \bigstrut\\
		\cline{2-4}          & $L_{24}$  & Linear & \tabincell{c}{In: $512$\\Out: $512$} \bigstrut\\
		\cline{2-4}          & $L_{25}$   & Linear & \tabincell{c}{In: $512$\\Out: $512$}  \bigstrut\\
		\cline{2-4}          & GRU   & GRU   & \tabincell{c}{In: $512$\\Out: $512$}  \bigstrut\\
		\cline{2-4}          & $\pi_{\bm{\theta}}$   & Linear & \tabincell{c}{In: $512$\\Out: $512$}  \bigstrut\\
		\hline
		FC & 1     & Linear & \tabincell{c}{In: $1024$\\Out: $2$}  \bigstrut\\
		\hline
	\end{tabular}%
	\label{tab-parameters}%
\end{table}%
\subsubsection{SiW} The Spoofing in the Wild (SiW for short) database \cite{FAS-Auxiliary-CVPR-2018} covers 165 subjects.
For each subject, eight genuine face videos and 20 spoofing face videos are recorded.
As for data collection environments, four sessions with variations of distances, poses, illuminations, and expressions have been considered \cite{FAS-Auxiliary-CVPR-2018}.
For print attack examples, an HP Color LaserJet M652 printer is for printing high resolution ($5184 \times 3456$) and low resolution ($1920 \times 1080$) photos. 
Besides, four e-devices (Samsung Galaxy S8, iPhone 7, iPad Pro, and PC ASUS MB168B) are used to collect spoofing faces on their screens.
As for cameras, a Canon EOS T6 and a Logitech C920 webcam are utilized to capture data.
Totally, the SiW database has up to 4478 genuine and spoofing face videos.
Also, it offers three protocols to evaluate the generalization capability of a model to unseen face poses and expressions (Protocol 1), unseen attack producing mediums (Protocol 2), and unseen Presentation Attack types (Protocol 3).

\subsubsection{ROSE-YOUTU}
The ROSE-YOUTU Face Liveness Detection Database (ROSE-YOUTU for short) is collected by the industry partner, YouTu. It involves 20 subjects, and for each subject, there are 25 genuine and 150 to 200 spoofing face videos. The data is diversely collected, covering up to 5 different lighting conditions.
Also, the front-facing cameras of five mobile devices (Hasee phone, Huawei phone, ZTE phone, iPad, iPhone 5s) were used to record the videos, with resolution ranging from $640 \times 480$ to $1280\times720$. 
Moreover, besides (still and quivering) printed photos and replay video examples (displayed with Lenovo LCD screen and Mac screen), the ROSE-YOUTU database further includes various paper mask attack examples. Such attacks can contain 3D information but are lacked in the aforementioned five databases. Hence, we leverage the ROSE-YOUTU database to evaluate our method further.
            
\subsection{Experiments Settings} \label{sec-exp-es}

\subsubsection{Evaluation protocols and metrics}
When evaluating the proposed framework, we report the experimental results in terms of Equal Error Rate (EER), Half-Total Error Rate (HTER), Attack Presentation Classification Error Rate (APCER), Bona Fide Presentation Classification Error Rate (BPCER), and Average Classification Error Rate (ACER) in different scenarios. 
For intra-database experiments on the CASIA, IDIAP, and ROSE-YOUTU databases, we use data of the training set of the given database to train models and report EER results on the corresponding testing sets.
When conducting cross-database experiments, we report HTER.
Besides, for the OULU-NPU and SiW databases, we respectively follow the four protocols \cite{OULU_NPU_2017} and the three protocols \cite{FAS-Auxiliary-CVPR-2018} to evaluate the generalization capability of our method by reporting ACER, APCER, and BPCER results. 


 
\subsubsection{Implementation Details}
 \textcolor{black}{As for the framework input, we consider including background information as spoofing-related clues are diverse and may not necessarily appear on face areas, which can be implemented by expanding face detection bounding boxes to crop out faces. However, there are various attack scenarios, and optimal bounding box sizes could depend on scenarios. Since the entire video frames can be regarded as the detected face cropped by the bounding box of a special configuration, we use such configuration by default for the framework input as a consistent way to evaluate the proposed framework. Nevertheless, we also evaluate the effectiveness of the framework under different configurations of bounding boxes, where the framework inputs are consistently resized to $256 \times 256$ pixels.}
Then, the backbone network embeds the input into feature maps \bm{$F$} $\in \mathbb{R}^{128\times32\times32}$. Subsequently, \bm{$F$} is forwarded to Branch 1 and Branch 2 to extract global and local features respectively.
For the final classification, we fuse the global and local features from Branch 1 and 2 by using the Concatenation as the input of the final Fully Connected (FC) layer.
We show the details of the backbone network, Branch 1, Branch 2, and the final FC layer in Table \ref{tab-parameters}, where $p$ denotes the size for cropping patches.


\begin{table}[tbp]
	\centering
	\caption{Performance comparisons between the models with only global features (Branch 1), with only local features (Branch 2) and with both fused on the CASIA, ROSE-YOUTU, REPLAY-ATTACK and the four protocols (P1, P2, P3, P4) of OULU-NPU databases. The performance is evaluated in terms of EER (\%). The \textbf{best results} are highlighted in bold. In the experiments, we set $p=8$ and $T=8$.}\label{tab-exp-ablation-drl}%
	\resizebox{\columnwidth}{!}{%
		\begin{tabular}{|l|c|c|c|}
			\hline
			Methods & Only local features & Only global features & Fused (ours) \bigstrut\\
			\hline
			CASIA & 3.26 & 0.66 & \textbf{0.17} \bigstrut\\
			\hline
			IDIAP & 2.54 & 0.88 & \textbf{0.00} \bigstrut\\
			\hline
			ROSE-YOUTU & 8.06 & 5.42 & \textbf{1.79} \bigstrut\\
			\hline
			OULU-NPU-P1 & 6.70 & 4.06 & \textbf{2.58} \bigstrut\\
			\hline
			OULU-NPU-P2 & 5.33 & 3.08 & \textbf{1.15}  \bigstrut\\
			\hline
			OULU-NPU-P3 & 3.38$\pm$3.78 & 2.21$\pm$1.48 & \textbf{1.18}$\pm$1.19 \bigstrut\\
			\hline
			OULU-NPU-P4 & 11.23$\pm$ 5.94 & 5.07$\pm$1.78 & \textbf{3.12}$\pm$2.01 \bigstrut\\
			\hline
		\end{tabular}%
	}
\end{table}%

\begin{figure}[tbp]
	\centering
	\includegraphics[width=1.0\linewidth]{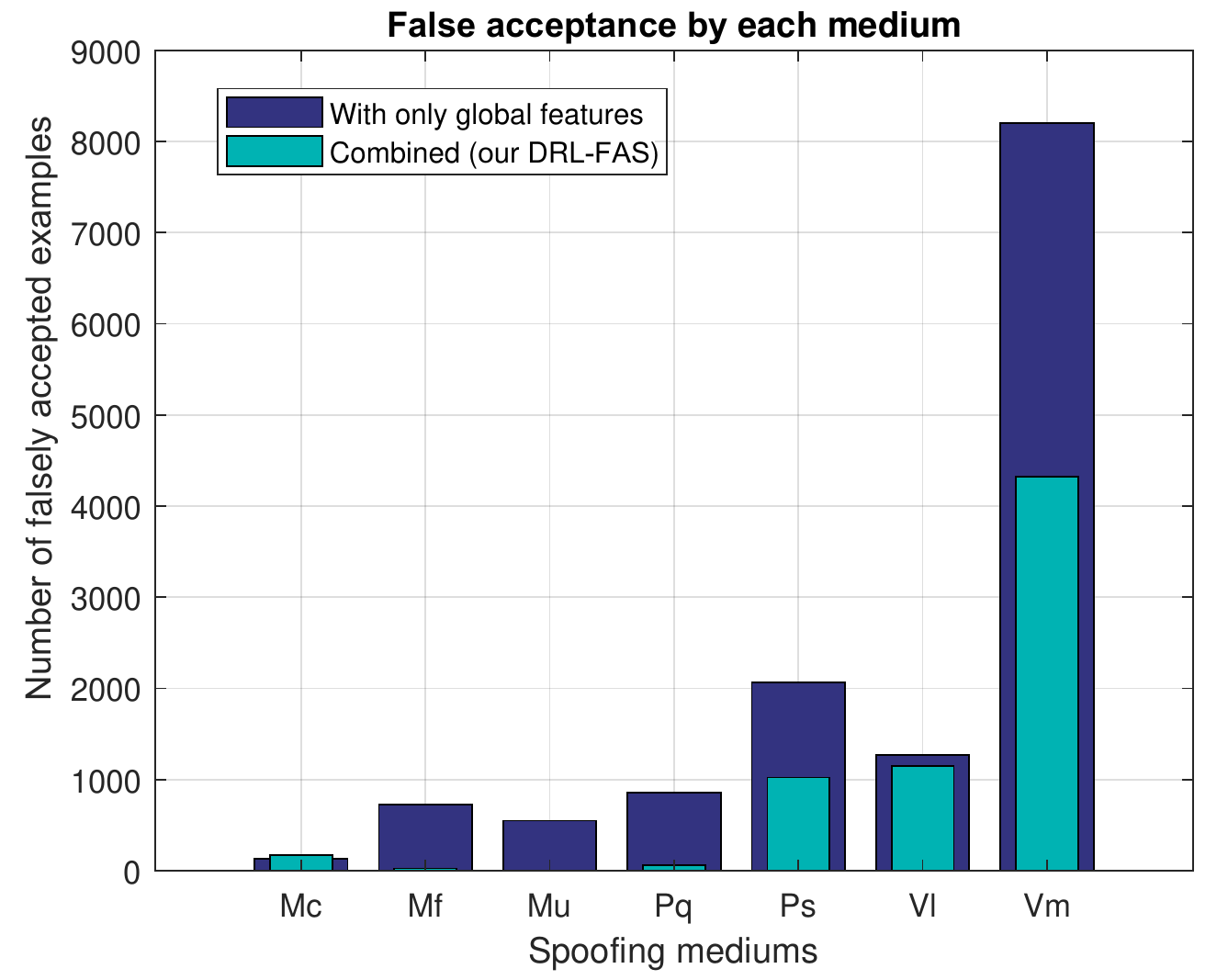}
	\caption{The statistics of the false acceptance examples on the ROSE-YOUTU database \cite{FAS-UnsupervisedDA-TIFS-2018}.
		The horizontal axis denotes the presentation attack mediums, and the vertical axis indicates the number of falsely accepted examples.
		``Mc'' indicates a paper mask with two eyes and mouth cropped out.
		``Mf'' indicates a paper mask without cropping.
		``Mu'' indicates a paper mask with the upper part cut in the middle.
		``Pq'' indicates a quivering printed paper.
		``Ps'' indicates a still printed paper.
		``Vl'' indicates a video that records a Lenovo LCD display. 
		``Vm'' indicates a video that records a Mac LCD display.
		Some examples of ``Vm" and ``Vl" can be seen in Fig. \ref{fig-examples}.
	}\label{fig-statistic}
\end{figure}

\par When training the framework, we follow the two-stage training scheme stated in Section \ref{sec-method}.
At the first stage, we pretrain a ResNet18 model \cite{CV-He-2016-ResNet} with cross-entropy loss with training data.
The trained parameters of the first convolutional layer and the four subsequent residual blocks are then loaded to the backbone, and the parameters of the backbone will be fixed and excluded from the second-stage optimization.
\textcolor{black}{At the second stage, the GRU's hidden state $\bm{h_t}$ is initialized as $\bm{f_g}$, and the location of the initial patch is sampled from a normal distribution whose symmetry center corresponds to the center of the input images.} Then, Branch 1 and 2 will be optimized jointly from scratch with the standard backward propagation with gradients of the cross-entropy loss and $J(\bm{\theta})$. By default, except for the declaration, the input configuration is ``FULL''; the number of observation steps $T$ is set as 8; the patch size $p$ is set as 8; and the fusion method is set as the Concatenation. In addition, we also explore the impacts of $p$, $T$, and different fusion methods in Section IV-C.

\subsection{Experimental results} \label{sec-exp-results}
\subsubsection{Analysis of Jointly Using Global and Local Features}\label{sec-exp-ablationstudy}
\par \textcolor{black}{
In this subsection, we demonstrate the effectiveness of our proposed framework by jointly utilizing global and local features. To this end, we ablate Branch 1 and Branch 2 separately to compare results with only local features and with only global features. As shown in Table~\ref{tab-exp-ablation-drl},  the results with only global features (Branch 2 ablated) are better than the results with only local features (Branch 1 ablated), which indicates that global features can be more effective than local features. Intuitively, people are likely to provide a reliable assessment with a glance at the original video frame, especially when discriminative artifacts appear, e.g., paper boundaries, bezels, or obvious reflections. However, when given merely a few local sub-patches, human beings could have trouble assessing the liveness of the original examples as discriminative artifacts may be absent in these patches. This is just like what the story \textit{Blind Men and An Elephant} \cite{book-The-Elephant-in-the-Dark} tells.  Moreover, by fusing the global and local features, our proposed framework can further achieve better performance. Such improvement supports our motivation that ``taking closer observations at local sub-patches'' can provide more information to refine the classification.
}
\par For further analysis, we collect the statistical results of the falsely accepted examples for each medium on the ROSE-YOUTU database.
Fig.~\ref{fig-statistic} shows that, with only global features, the number of falsely-accepted ``Vm" (video attack recorded from a Mac display) is nearly 9000, which represents the largest proportion. 
By carefully reviewing the data in the ROSE-YOUTU database, we find that the visual quality of ``Vm'' is generally better than that of ``Vl'' (video attack recorded from a Lenovo display) as the resolution of a Mac display ($2560 \times 1600$ resolution) is much higher than that of a Lenovo LCD ($1920 \times 1080$ resolution).
For instance, Fig.~\ref{fig-examples}(c) and is a ``Vl'' example, and Fig.~\ref{fig-examples}(e) is a ``Vm'' example. As shown, the visual quality of the ``Vm" looks better and than the ``Vl''.
In other words, the spoofing faces of ``Vm" visually look more similar to genuine faces than ``Vl'' (as well as others). Therefore, ``Vm" examples get most falsely accepted as genuine faces than ``Vl'' and the others.
Moreover, as shown in Fig.~\ref{fig-statistic}, our method can generally lead to fewer falsely accepted spoofing examples of various attack mediums.
Although the falsely accepted ``Vm" from our method still accounts for the largest proportion of false acceptance, the value is nearly half of that with only global features.
This means that our method can better discriminate spoofing examples of good visual quality by leveraging local information from sub-patches.
It corresponds to our motivation that people can zoom in local sub-patches and explore subtle spoofing clues to refine and confirm their assessment of the liveness of examples.
\textcolor{black}{By far, we demonstrate that our framework can better counter spoofing attacks by jointly using global and local features. In the next subsection, we investigate how well our framework can be generalized to inputs of other configurations.}

\begin{figure}
	\centering
	\subfigure[]{
		\label{fig-1a}
		\includegraphics[width=0.16\linewidth]{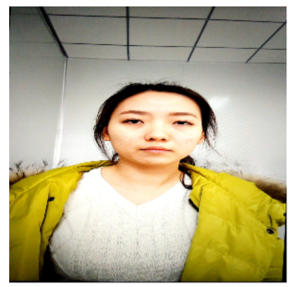}
	}
	\subfigure[]{
		\includegraphics[width=0.16\linewidth]{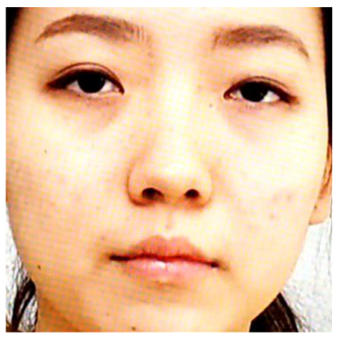}
	}
	\subfigure[]{
		\includegraphics[width=0.16\linewidth]{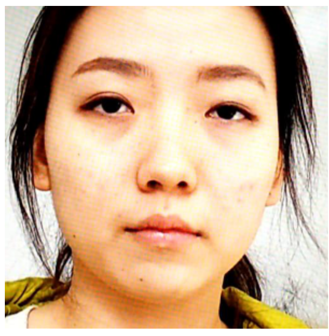}
	}
	\subfigure[]{
		\includegraphics[width=0.16\linewidth]{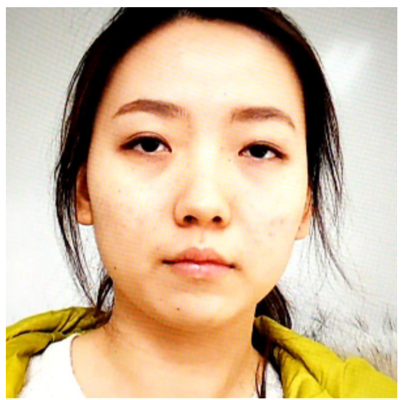}
	}
	\subfigure[]{
		\includegraphics[width=0.16\linewidth]{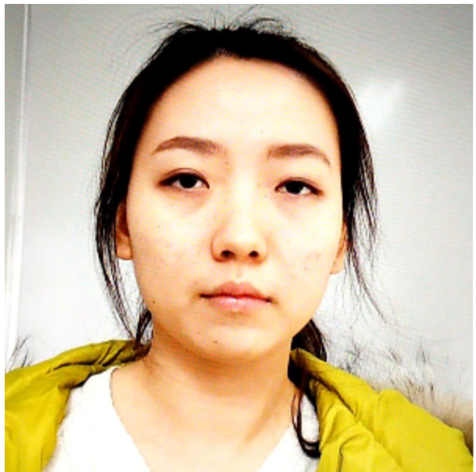}
	}
	
	\caption{Illustrations of the faces cropped by bounding boxes of different configurations. (a) is the entire original video frame of a spoofing example from the OULU-NPU database, which can be regarded as the face image cropped by a bounding box with a special configuration (``FULL''). (b) is the detected face cropped from (a) by a bounding box with the default setting of the dlib's CNN detector ($\alpha=0.0$). Analogously, (c) is that by 20\% ($\alpha=0.2$), (d) is that by 40\% ($\alpha=0.4$), and (e) is that by 60\% ($\alpha=0.6$).}

	\label{fig-ext-examples}
\end{figure} 
\begin{table}[t]
	\centering
	\caption{The comparison between the ResNet18 baseline and our proposed framework. The ACER (\%) results are reported on the four protocols (P1, P2, P3, P4) of the OULU-NPU database. The \textbf{best results} are highlighted in bold. In the experiments, we set $p=8$ and $T=8$.}
	\begin{tabular}{|l|M{1cm}|M{1cm}|M{1cm}|M{1cm}|}
		\hline
		\multirow{2}[4]{*}{Method} & \multicolumn{4}{c|}{ACER} \bigstrut\\
		\cline{2-5}          & P1    & P2    & P3    & P4 \bigstrut\\
		\hline
		ResNet18 ($\alpha=0.0$) & 4.0   & 4.5   & 7.2$\pm$7.0 & 10.4$\pm$6.0 \bigstrut\\
		\hline
		ResNet18 ($\alpha=0.2$) & 5.3   & 3.3   & 4.8$\pm$5.0 & 9.4$\pm$7.1 \bigstrut\\
		\hline
		ResNet18 ($\alpha=0.4$) & 2.4   & 3.6   & 4.5$\pm$4.1 & 7.4$\pm$3.8 \bigstrut\\
		\hline
		ResNet18 ($\alpha=0.6$) & 2.3   & 3.7   & 5.1$\pm$2.4 & 6.3$\pm$2.6 \bigstrut\\
		\hline
		ResNet18 (FULL) & 7.2   & 2.3   & 3.4$\pm$1.3 & 11.4$\pm$5.1 \bigstrut\\
		\hline
		Ours ($\alpha=0.0$) & 3.7   & 4.0   & 6.4$\pm$6.9 & 9.5$\pm$6.6 \bigstrut\\
		\hline
		Ours ($\alpha=0.2$) & 4.4   & 3.0   & 3.9$\pm$2.3 & 8.6$\pm$5.6 \bigstrut\\
		\hline
		Ours ($\alpha=0.4$) & 1.6   & 3.4   & \textbf{2.9}$\pm$1.4 & 6.0$\pm$4.6 \bigstrut\\
		\hline
		Ours ($\alpha=0.6$) & \textbf{1.4}   & 2.6   & 3.1$\pm$1.8 & \textbf{5.0}$\pm$3.7 \bigstrut\\
		\hline
		Ours (FULL) & 4.7   & \textbf{1.9}   & 3.0$\pm$1.5 & 7.2$\pm$3.9 \bigstrut\\
		\hline
	\end{tabular}%
	\label{tab-oulu-ext}%
\end{table}%


\begin{table}[htbp]
	\centering
	\caption{The comparison of different methods for selecting patches. The performance is evaluated in terms of EER (\%). The \textbf{best results} are highlighted in bold. In the experiments, we set $p=8$ and $T=8$.}
	\begin{tabular}{|l|c|c|c|}
		\hline
		EER & MAX-SCORES & RANDOM & DRL (ours) \bigstrut\\
		\hline
		CASIA & 0.25 &0.32 & \textbf{0.17} \bigstrut\\
		\hline
		IDIAP & 0.00 & 0.00 & \textbf{0.00} \bigstrut\\
		\hline
		ROSE-YOUTU & 3.29 & 2.14  & \textbf{1.79} \bigstrut\\
		\hline
		OULU-NPU-P1 & 3.82 &3.57  & \textbf{2.58} \bigstrut\\
		\hline
		OULU-NPU-P2 & 1.54 &1.76  & \textbf{1.15}\bigstrut\\
		\hline
		OULU-NPU-P3 & 1.20$\pm$1.03 & 1.38$\pm$0.94 & \textbf{1.18}$\pm$1.12 \bigstrut\\
		\hline
		OULU-NPU-P4 & 4.67 $\pm$1.31  &4.32$\pm$1.75 & \textbf{3.12}$\pm$2.01 \bigstrut\\
		\hline
	\end{tabular}%
	\label{tab-r2-selection}%
\end{table}%

\subsubsection{Analysis of Configurations of the Framework Input}
\par \textcolor{black}{As our framework is proposed to exploit the discriminative information which may not necessarily appear on face areas, we also propose to investigate the performance by configuring the input with different scales of information from backgrounds based on detected faces.
    To this end,  we propose to use a \textit{dlib}'s CNN face detector \cite{dlib09} to obtain detection bounding boxes. Subsequently, a bounding box will be expanded by 0\%, 20\%, 40\%, and 60\% (i.e., $\alpha=\{0.0, 0.2, 0.4, 0.6\}$) to produce four face images that have different scales of background information. Besides, using the entire video frames can be treated as cropping faces with a special configuration for the bounding box, and we denote such configuration as ``FULL''. As such, there will be five groups of face images in different configurations for the framework input, and some of the examples are shown in Fig.~\ref{fig-ext-examples}. For experiments, we adopt the OULU-NPU database, as it provides four protocols for extensive evaluation. Also, we train ResNet18 models to provide the baseline results, where only global features are considered. The experimental results are shown in Table~\ref{tab-oulu-ext}.
    It is obvious that our method can achieve better ACER results than the counterparts of ResNet18 over different input configurations. To sum up, the experiments show that the proposed method can still be effective in jointly using global and local features extracted from face images that have different scales of background information.
   }

\subsubsection{Effect of using reinforcement learning}
\textcolor{black}{In this subsection, we show the effectiveness of adopting deep reinforcement learning (DRL) for selecting patches.
        \textcolor{black}{To be more specific, we compare our proposed DRL with the method of selecting patches that have the max SoftMax scores (denoted as the MAX-SCORES method), and the method of selecting patches randomly (denoted as the RANDOM method). In the implementation of the MAX-SCORES method, we pretrained a patch-based CNN based on \cite{FAS-Depth&Patch-IJCB-2018} with the training data to infer the SoftMax scores of the patches, and those patches that have the max SoftMax scores are selected for the framework.
        Besides, in the implementation of the RANDOM method, we use a random number generator to generate locations of the patches to be selected.
        Table~\ref{tab-r2-selection} compares the performance of the MAX-SCORES, RANDOM, and the DRL with respect to patch selection.}
        Regarding the IDIAP database, the MAX-SCORES, RANDOM, and DRL methods achieve 0.00\% EER, meaning that local features can help with the final prediction, regardless of how we select patches. Nevertheless, in the other experiments based on the ROSE-YOUTU, CASIA, and OULU-NPU databases, our proposed DRL shows the effectiveness in selecting patches by achieving the best EER results.  
        }

\begin{table}[tbp]
	\centering
	\caption{The EER results of different patch size $p$ on the CASIA, ROSE-YOUTU and Protocol 1 of the OULU-NPU database (OULU-NPU-P1). The \textbf{best results} are highlighted in bold, and the \underline{second-best results} are underlined. In the experiments, we set $T=8$. }
	\begin{tabular}{|l|c|c|c|}
		\hline
		EER     & CASIA & ROSE-YOUTU  & OULU-NPU-P1 \bigstrut\\
		\hline
		$p=2$    & 0.35  & 3.92  & \textbf{2.29} \bigstrut\\
		\hline
		$p=4$     & \textbf{0.17}  & \underline{2.73}  & 2.95 \bigstrut\\
		\hline
		$p=8$     & \textbf{0.17}  & \textbf{1.79}  & \underline{2.58} \bigstrut\\
		\hline
		$p=16$    & \underline{0.28}  & 3.65   & 4.20 \bigstrut\\
		\hline
	\end{tabular}%
	\label{tab-p}%
\end{table}%
\subsubsection{Analysis of Local Features Extraction}
    In this subsection, we analyze the impact of patch size $p$ and the number of steps $T$ for the local feature extraction.
\par \textbf{Effect of patch size $p$}  
\textcolor{black}{To analyze the effect of $p$, we conduct experiments with $p=\{2, 4, 8, 16\}$, and we provide the EER results on the CASIA, ROSE-YOUTU, and the OULU-NPU-P1 in Table~\ref{tab-p}. Regarding the CASIA database, we can see that the EER results become better from 0.35\% to 0.17\% when $p$ increases from 2 to 8 but deteriorate when we further increase $p$ to $16$ (0.28\%).
                Regarding the ROSE-YOUTU database, a similar trend can be observed that the EER performance improves up to 1.79\% EER when $p$ increases from 2 to 8. However, when $p=16$, the EER performance drops as 3.65\% EER. Regarding the OULU-NPU-P1, the best EER is achieved when $p=2$, and the EER result of $p=8$ is better than that of $p=4$ and $p=16$.
                Therefore, the optimal $p$ is different for different databases, and simply increasing $p$ may not necessarily lead to better performance. Nevertheless, we observe that $p=8$ can achieve the desired performance in general, and thus we fix $p=8$ for all other experiments.}
                
\begin{table}[t]
	\centering
	\caption{The EER results of different total number of observation steps $T$ on the CASIA, ROSE-YOUTU, and Protocol 1 of the OULU-NPU database (OULU-NPU-P1). The \textbf{best results} are highlighted in bold, and the \underline{second-best results} are underlined. In the experiments, we set $p=8$.}
	\begin{tabular}{|l|c|c|c|}
		\hline
		EER & CASIA & ROSE-YOUTU  & OULU-NPU-P1 \bigstrut\\
		\hline
		$T=2$     & 0.18  & \textbf{1.64}  & 3.85 \bigstrut\\
		\hline
		$T=4$     & \underline{0.17}  & 2.13  & \underline{3.15} \bigstrut\\
		\hline
		$T=8$     & \underline{0.17}  & \underline{1.79}  & \textbf{2.58} \bigstrut\\
		\hline
		$T=16$    & \textbf{0.06}  & 1.92  & 3.52 \bigstrut[t]\\
		\hline
	\end{tabular}%
	\label{tab-T}%
\end{table}%

\par \textbf{Effect of total number of observation steps $T$}
\textcolor{black}{
                Besides the size of the local patches, we are also interested in the impact of different numbers of observation steps $T$.
                To this end, we conduct experiments by increasing $T$ from 2 to 16, and the results are reported in Table~\ref{tab-T}. As we can observe, for the CASIA database, the EER performance improves when $T$ increases from 2 to 16.
                Regarding the ROSE-YOUTU database, the best EER result of 1.65\% is achieved when $T=2$, and the second-best EER of 1.79\% is achieved when $T=8$. Regarding the OULU-NPU-P1, the EER performance improves as $T$ changes from 2 to 8 but gets worse when $T=16$. 
                In summary, simply increasing $T$ does not necessarily lead to better performance.
                We conjecture the reason that when $T$ is increased to include more patches, those patches containing less discriminative information may deteriorate the performance. Nevertheless, we consistently choose $T=8$ for other experiments as it can lead to the best or second-best performance in Table~\ref{tab-T}.    
            } 

\begin{table}[t]
	\centering
	\caption{The comparison between with/without the Backbone for the feature embedding. The experiments are conducted on the CASIA, ROSE-YOUTU, REPLAY-ATTACK, OULU-NPU databases. The performance is evaluated in terms of EER (\%). The \textbf{best results} are highlighted in bold. In the experiments, we set $p=8$ and $T=8$.}
	
	\begin{tabular}{|l|c|c|}
		\hline
		Methods & \multicolumn{1}{l|}{Without backbone } & \multicolumn{1}{l|}{With backbone} \bigstrut\\
		\hline
		CASIA & 7.80 & \textbf{0.17} \bigstrut\\
		\hline
		IDIAP & 1.13 & \textbf{0.00} \bigstrut\\
		\hline
		ROSE-YOUTU & 2.38 & \textbf{1.79} \bigstrut\\
		\hline
		OULU-NPU-P1 & 8.43  & \textbf{2.58} \bigstrut\\
		\hline
		OULU-NPU-P2 & \textbf{1.04}  &  1.15 \bigstrut\\
		\hline
		OULU-NPU-P3 & \textbf{1.15} $\pm$1.26 & 1.18$\pm$1.19 \bigstrut\\
		\hline
		OULU-NPU-P4 & 9.57$\pm$5.35 &  \textbf{3.12}$\pm$2.01\bigstrut\\
		\hline
	\end{tabular}%
	\label{tab-exp-backbone}
\end{table}%

\begin{table}[t]
	\centering
	\caption{Performance comparisons between the one-stage end-to-end training and our two-stage training on the CASIA database. The performance is evaluated  in terms of EER (\%).  In the experiments, we set $p=8$. }\label{tab-two-stage-training}
	\begin{tabular}{|p{2cm}<\centering|p{2cm}<\centering|p{2cm}<\centering|}
		\hline
		Observation steps & One-stage & Our two-stage \bigstrut\\
		\hline
		$T=2$   & 20.1  & 0.184 \bigstrut\\
		\hline
		$T=4$  & 18.7  & 0.171 \bigstrut\\
		\hline
		$T=8$  & 4.32  & 0.171 \bigstrut\\
		\hline
	\end{tabular}%
\end{table}%

 \subsubsection{Analysis of other settings of the framework}
    In this subsection, we analyze the impacts of the backbone, training strategies, and feature fusion methods in our proposed framework.
\par \textbf{Effect of using the backbone for local features}
\textcolor{black}{To study the effectiveness of using the backbone for feature embedding and local feature extraction, we also implement the framework without the backbone for feature embedding as a baseline, where patches are cropped at the predicted locations from original RGB inputs instead of the embedded feature maps.
Table~\ref{tab-exp-backbone} shows the results for comparison.
We observe that better performances can be achieved in general by considering a backbone network. Such improvement indicates that extracting local features from the feature maps through the backbone can help to extract more discriminative spoofing-related information. Moreover, the results with the backbone on the OULU-NPU-P1 and -P4 (2.58\% and 3.12\%) are significantly better than those without the backbone (8.43\% and 9.57\%). As Protocol 1 and 4 involve unseen environments in the testing data, conducting feature embedding through the backbone network to extract local features may alleviate environmental interference to some extent such that better performance can be achieved.}

\textbf{Effect of the two-stage training} Although training our framework in an end-to-end manner is achievable, we propose to use a two-stage training scheme to better optimize our framework.
Table \ref{tab-two-stage-training} compares the results of the one-stage end-to-end training and our proposed two-stage training.
The EER results of our two-stage training are all remarkably lower than 0.2\% for $T=2, 4, 8$.
However, for the one-stage training experiments, when $T=2$, the EER result is up to 20.1\%.
Although the EER decreases as $T$ increases, the best EER result is still above 4\% (${T=8}$), which is much higher than all the results achieved by our two-stage training.
Therefore, Table \ref{tab-two-stage-training} shows that our two-stage training can help achieve better results by providing a stable environment such that the agent can learn to extract effective features, even when ${T=2}$.

\begin{table}[tbp]
	\footnotesize
	\begin{center}
		\caption{The results on the CASIA database in different settings of fusion methods. The performance is evaluated in terms of EER~(\%).} \label{tab-exp-fusion}%
		\resizebox{\columnwidth}{!}{%
			\begin{tabular}{|c|c|c|c|c|}
				\hline
				Observation steps & Patch size & Average & Weighted Average & Concatenation  \\
				\hline
				\multirow{2}*{$T=4$}   & $p=4$    & 16.1	& 12.3&	0.184     \\  
				\cline{2-5}
				&  $p=8$  &   9.54	& 11.6&	0.171 \\ \hline
				
				\multirow{2}*{$T=8$}   & $p=4$   & 15.6&	12.4 &	0.171\\ 
				\cline{2-5}
				&  $p=8$ & 0.132 & 0.172 & 0.171  \\ \hline    		
			\end{tabular}
		}
	\end{center}
	\hspace{1cm}
	\setlength{\textfloatsep}{0.1cm}
\end{table}

\textbf{Effect of fusion methods}
In this framework, the global and local features are fused for classification. In Table~\ref{tab-exp-fusion}, we compare results among three different fusion methods, the Average, Weighted Average \cite{FAS-MSR-TIFS-2019}, and Concatenation.
We observe that the EER results are all lower than 0.2\% when $T=8$ and $p=8$.  
However, when $T<8$ or $p<8$, the EER results of the Average and the Weighted Average are higher than 9\%, while the result with only global features is less than 1\% (shown in Table \ref{tab-exp-ablation-drl}).
We conjecture the reason that the Average and Weighted Average fuse global and local features by averaging the elements at each dimension. When $T<8$ or $p<8$, the extracted local features may not be effective enough. As a result, discriminative information contained by global features may be distorted after the average operation. However, when the extracted local feature is not representative enough, the Concatenation could maintain original global information to a larger extent. 
Therefore, the Concatenation can provide stable results (all lower than 0.2\% EER) under different settings of $T$ and $p$, and we fix the Concatenation as the fusion method in other experiments.

\begin{table}[tbp]
	\centering
	\caption{Intra-database experiments on the CASIA database and the IDAIP database. The performance is evaluated in terms of EER (\%) and HTER (\%). ``$-$'' means the result is not available. The \textbf{best results} are highlighted in bold In the experiments, we set $p=8$ and $T=8$.}
	\begin{tabular}{|p{11 em}<\centering|c|c|c|}
		\hline
		\multirow{2}[4]{*}{Method} & \multicolumn{2}{p{8.43em}<\centering|}{IDIAP} & \multicolumn{1}{p{4.215em}<\centering|}{CASIA} \bigstrut\\
		\cline{2-4}    \multicolumn{1}{|c|}{} & \multicolumn{1}{p{4.215em}<\centering|}{EER} & \multicolumn{1}{p{4.215em}<\centering|}{HTER} & \multicolumn{1}{p{4.215em}<\centering|}{EER} \bigstrut\\
		\hline
		\hline
		CNN \cite{FAS-CNN-ComputerScience-2014}  & 6.1   & 2.1   & 7.4 \bigstrut\\
		\hline
		Color-LBP \cite{FAS-ColorTexture-TIFS-2016}  & 0.4   & 2.9   & 6.2 \bigstrut\\
		\hline
		Bottleneck feature\newline{}fusion + NN \cite{BottleCNN-JVCIR-2016}  & 0.8  & \textbf{0.0} & 5.8 \bigstrut\\
		\hline
		MSR-ResNet \cite{FAS-MSR-TIFS-2019} & 0.2  & 0.4 & 3.1 \bigstrut\\
		\hline
		DRL-FAS (ours)  & \textbf{0.0} & \textbf{0.0} & \textbf{0.2} \bigstrut\\
		\hline
	\end{tabular}%
	\label{tab-intra-casia-idiap}%
\end{table}%

\begin{table}[t]
	\centering
	\caption{The intra-database experiment results on the ROSE-YOUTU database compared with the state-of-the-art methods. The performance is evaluated in terms of EER (\%). The \textbf{best result} is highlighted in bold. In the experiments, we set $p=8$ and $T=8$.}
	\begin{tabular}{|p{4cm}<\centering|p{2cm}<\centering|}
		\hline
		\multicolumn{1}{|c|}{\multirow{2}[4]{*}{Method}} & ROSE-YOUTU \bigstrut\\
		\cline{2-2}          & EER \bigstrut\\
		\hline
		CoALBP (YCBCR) \cite{FAS-CoALBP-AIVT-2012P} & 17.1 \bigstrut\\
		\hline
		CoALBP (HSV) \cite{FAS-CoALBP-AIVT-2012P} & 16.4 \bigstrut\\
		\hline
		AlexNet \cite{FAS-UnsupervisedDA-TIFS-2018} & 8.0 \bigstrut\\
		\hline
		3D-CNN \cite{FAS-3DCNN-TIFS-2018}& 7.0 \bigstrut\\
		\hline
		DeSpoofing \cite{FAS-DeSpoofing-ECCV-2018} & 12.3\bigstrut\\
		\hline
		DRL-FAS (ours)   & \textbf{1.8} \bigstrut\\
		\hline
	\end{tabular}%
	\label{tab-rose-intra}%
\end{table}%


\subsubsection{Intra-database experiments}
For further evaluation, we conduct experiments on six benchmark databases and compare our proposed method with state-of-the-art methods.

\par \textbf{Results on the CASIA database and IDIAP database}
Table \ref{tab-intra-casia-idiap} provides the results of intra-database experiments on the CASIA and IDIAP databases.
On the CASIA database, our method attains 0.17\% EER, which is the best.
The best performance on the IDIAP database can also be seen from the 0\% EER on the development (DEV) set and the 0\% HTER on the testing (TEST) set.
On both the two benchmark databases, our method achieves the best performance and shows its effectiveness.
\begin{table}[tbp]
	\centering
	\footnotesize
	\caption{Comparison between the proposed framework and
		state-of-the-art methods on the SiW database. The performance is evaluated in terms of ACER (\%). The \textbf{best results} are highlighted in bold. In the experiments, we set $T=8$ and $p=8$.} \label{tab-exp-siw}%
	\begin{threeparttable}
		\begin{center}	
			\begin{tabular}{|p{1cm}<\centering|p{2cm}<\centering|p{2cm}<\centering|}
				\hline
				Protocol            & Method    & ACER \\ \hline	
				\multirow{4}*{1} & Auxiliary \cite{FAS-Auxiliary-CVPR-2018} & 1.00	\\\cline{2-3}
				& STASN \cite{FAS-CVPR2019-STASN}     & 1.00 \\\cline{2-3}
				& STASN+ \cite{FAS-CVPR2019-STASN}      & 0.30 \\\cline{2-3}
				& DRL-FAS (ours)                           & \textbf{0.00}\\\hline \hline
				
				\multirow{4}*{2} & Auxiliary \cite{FAS-Auxiliary-CVPR-2018} & 0.57$\pm$0.69 \\\cline{2-3}
				& STASN \cite{FAS-CVPR2019-STASN}       & 0.28$\pm$0.05 \\\cline{2-3}
				& STASN+ \cite{FAS-CVPR2019-STASN}      & 0.15$\pm$0.05 \\\cline{2-3}
				& DRL-FAS (ours)      & \textbf{0.00}$\pm$0.00 \\ \hline \hline
				
				\multirow{4}*{3} & Auxiliary \cite{FAS-Auxiliary-CVPR-2018} & 8.31$\pm$3.81 \\\cline{2-3}
				& STASN \cite{FAS-CVPR2019-STASN}       & 12.10$\pm$1.50 \\\cline{2-3}
				& STASN+ \cite{FAS-CVPR2019-STASN}      & 5.85$\pm$0.85 \\\cline{2-3}
				& DRL-FAS (ours)       & \textbf{4.51}$\pm$0.00\\\hline
				
			\end{tabular}	
			\begin{tablenotes}
				\item $\bullet$ Protocol 1, 2, and 3 are for evaluating models' generalization capability to unseen face poses and expressions, unseen attack mediums, and unseen Presentation Attack types, respectively.
				\item $\bullet$ For experiments of Protocol 1, the testing is only done once so there are no terms of standard deviation
			\end{tablenotes}
		\end{center}
	\end{threeparttable}
\end{table}

\begin{table*}[htbp]
	\centering
	\begin{adjustbox}{width=0.95\textwidth} 
		\begin{threeparttable}
			\caption{Experiment results for the four protocols on the OULU-NPU database. The performance is evaluated in terms of APCER (\%), BPCER (\%) and ACER (\%) on the testing ({Test}) set. In the experiments, we set $p=8$ and $T=8$. The \textbf{best results} are highlighted in bold.} \label{tab-exp-oulu}
			\begin{tabular}{|c|c|c|c|c|c|c|c|c|c|}
				\hline
				\multicolumn{1}{|c|}{\multirow{2}[4]{*}{Protocol}} & \multirow{2}[4]{*}{Method}   & \multicolumn{3}{c|}{\centering Test} & \multicolumn{1}{c|}{\multirow{2}[4]{*}{Protocol}} & \multirow{2}[4]{*}{Method }  & \multicolumn{3}{c|}{Test} \bigstrut\\
				\cline{3-5}\cline{8-10}        & \multicolumn{1}{c|}{}   & APCER & BPCER & ACER  &    &   \multicolumn{1}{c|}{}  & APCER  & BPCER & ACER \bigstrut\\
				\hline
				\cline{2-5}\cline{7-10} \multirow{4}[14]{*}{1}       & GRADANT \cite{FAS-Competetion-IJCB-2017}  & \multicolumn{1}{c|}{1.3} & \multicolumn{1}{c|}{12.5} & \multicolumn{1}{c|}{6.9} &    \multirow{4}[14]{*}{2}  &   GRADANT\cite{FAS-Competetion-IJCB-2017}  &  \multicolumn{1}{c|}{3.1} & \multicolumn{1}{c|}{1.9} & \multicolumn{1}{c|}{2.5} \bigstrut\\
				\cline{2-5}\cline{7-10}          & DeSpoofing\cite{FAS-DeSpoofing-ECCV-2018}  & \multicolumn{1}{c|}{\textbf{1.2}} & \multicolumn{1}{c|}{1.7} & \multicolumn{1}{c|}{\textbf{1.5}} &       & DeSpoofing\cite{FAS-DeSpoofing-ECCV-2018}   & \multicolumn{1}{c|}{4.2} & \multicolumn{1}{c|}{4.4} & \multicolumn{1}{c|}{4.3} \bigstrut\\
				\cline{2-5}\cline{7-10}  & Auxiliary \cite{FAS-Auxiliary-CVPR-2018} & \multicolumn{1}{c|}{1.6} & \multicolumn{1}{c|}{\textbf{1.6}} & \multicolumn{1}{c|}{1.6} &       & Auxiliary \cite{FAS-Auxiliary-CVPR-2018} & \multicolumn{1}{c|}{\textbf{2.7}} & \multicolumn{1}{c|}{2.7} & \multicolumn{1}{c|}{2.7} \bigstrut\\
				\cline{2-5}\cline{7-10}          & MSR-ResNet \cite{FAS-MSR-TIFS-2019} &  \multicolumn{1}{c|}{5.1} & \multicolumn{1}{c|}{6.7} & \multicolumn{1}{c|}{5.9} &       & MSR-ResNet \cite{FAS-MSR-TIFS-2019} & \multicolumn{1}{c|}{7.6} & \multicolumn{1}{c|}{2.2} & \multicolumn{1}{c|}{4.9} \bigstrut\\
				\cline{2-5}\cline{7-10}          & DRL-FAS (ours)  & \multicolumn{1}{c|}{5.4} & \multicolumn{1}{c|}{4.0} & \multicolumn{1}{c|}{4.7} &      & DRL-FAS (ours)  & \multicolumn{1}{c|}{3.7} & \multicolumn{1}{c|}{\textbf{0.1}} & \multicolumn{1}{c|}{\textbf{1.9}} \bigstrut\\
				\hline
				\cline{1-10} \multirow{4}[14]{*}{3}       & GRADANT \cite{FAS-Competetion-IJCB-2017}  &  \textbf{2.6}$\pm$3.9  & 5.0$\pm$5.3  & 3.8$\pm$2.4 &  \multirow{4}[14]{*}{4} &    GRADANT \cite{FAS-Competetion-IJCB-2017}  &  \textbf{5.0}$\pm$4.5  & 15.0$\pm$7.1  & 10.0$\pm$5.0 \bigstrut\\
				\cline{2-5}\cline{7-10}          & DeSpoofing \cite{FAS-DeSpoofing-ECCV-2018}  &  4.0$\pm$1.8  & 3.8$\pm$1.2  & 3.6$\pm$1.6 &    &   DeSpoofing \cite{FAS-DeSpoofing-ECCV-2018}  &  5.1$\pm$6.3  & \textbf{6.1}$\pm$5.1  & \textbf{5.6}$\pm$5.7 \bigstrut\\
				\cline{2-5}\cline{7-10}   & Auxiliary \cite{FAS-Auxiliary-CVPR-2018}   & 2.7$\pm$1.3  & 3.1$\pm$1.7  & \textbf{2.9}$\pm$1.5 &       & Auxiliary \cite{FAS-Auxiliary-CVPR-2018}  & 9.3$\pm$5.6  & 10.4$\pm$6.0  & 9.5$\pm$6.0 \bigstrut\\
				\cline{2-5}\cline{7-10}          & MSR-ResNet \cite{FAS-MSR-TIFS-2019} &  3.9$\pm$2.8  & 7.3$\pm$1.1  & 5.6$\pm$1.6 &    &   MSR-ResNet \cite{FAS-MSR-TIFS-2019} &  11.3$\pm$3.9  & 9.7$\pm$4.8  & 9.8$\pm$4.2 \bigstrut\\
				\cline{2-5}\cline{7-10}    & DRL-FAS (ours) 
				& 4.6$\pm$3.6  & \textbf{1.3}$\pm$1.8  & 3.0$\pm$1.5 &   & DRL-FAS (ours, FULL)    
				& 8.1$\pm$2.7 &  6.9$\pm$5.8 &  7.2$\pm$3.9 \bigstrut\\
				\hline
			\end{tabular}    
			
			\begin{tablenotes}
				\item $\bullet$ Protocol 1, 2, and 3 are for evaluating a model's generalization capability to unseen environment conditions, unseen attack mediums, and unseen camera modules, respectively. Protocol 4 extends the evaluation to unseen sessions, attack mediums and camera modules at the same time.
				\item $\bullet$ For experiments of Protocol 1 and 2, the testing is only done once, so there are no terms of standard deviation.
			\end{tablenotes}
		\end{threeparttable}
	\end{adjustbox}
	
\end{table*}%

\par \textbf{Results on the ROSE-YOUTU database}
Table \ref{tab-rose-intra} compares our method with the baseline methods on the ROSE-YOUTU database. Our method can achieve the lowest EER (1.8\%), while the second-best method 3D-CNN \cite{FAS-3DCNN-TIFS-2018} merely achieves 7.0\% EER. This shows our method's superiority. 
In addition, to evaluate how paper mask attacks in the ROSE-YOUTU database can fail depth-based methods, we implement the DeSpoofing method \footnote{https://github.com/yaojieliu/ECCV2018-FaceDeSpoofing} \cite{FAS-DeSpoofing-ECCV-2018} because it leverages depth information for training. 
However, it merely achieves 12.3\% EER, which indicates that when encountering paper mask attack examples that are with depth information, depth-based methods could lose efficacy. By contrast, our method can still perform favorably against the paper mask attack examples in the ROSE-YOUTU database.

\begin{table*}[tbp]
	\centering
	\caption{Inter-database results between the CASIA, IDIAP and MSU databases. The performance is evaluated in terms of HTER (\%). ``$-$'' means the result is not available. On the left of ``$\rightarrow$" is the database used for training and on the right for testing. We mark the best results in bold, and the second-best results are underlined. In the experiments, we set $p=8$ and $T=8$.}  \label{tab-exp-cross-casia-msu-idiap}  
	\begin{tabular}{|c|c|c|c|c|c|}
		\hline
		Method & CASIA$\rightarrow$IDIAP & IDIAP$\rightarrow$CASIA  & MSU$\rightarrow$IDIAP & IDIAP$\rightarrow$MSU \bigstrut\\
		\hline
		LBP \cite{FAS-CanFace-ICB-2013}  & 47.0  & 39.6  & 45.5  & 45.8  \bigstrut\\
		\hline
		LBP-TOP \cite{FAS-CanFace-ICB-2013}  & 49.7  & 60.6  & 46.5  & 47.5  \bigstrut\\
		\hline
		Motion \cite{FAS-CanFace-ICB-2013} & 50.2  & 47.9  & \multicolumn{1}{c|}{$-$} & \multicolumn{1}{c|}{$-$} \bigstrut\\
		\hline
		CNN \cite{FAS-CNN-ComputerScience-2014}  & 48.5  & 45.5  & 37.1  & 48.6  \bigstrut\\
		\hline
		Color LBP \cite{FAS-ColorTexture-TIFS-2016}  & 37.9  & 35.4  & 44.8  & 33.0  \bigstrut\\
		\hline
		Color Texture \cite{FAS-ColorTexture-TIFS-2016}  & 30.3  & 37.7  & 33.9  & 34.1  \bigstrut\\
		\hline
		Auxiliary \cite{FAS-Auxiliary-CVPR-2018}  & \textbf{27.6}  & \textbf{28.4}  & \multicolumn{1}{c|}{$-$} & \multicolumn{1}{c|}{$-$} \bigstrut\\
		\hline
		DeSpoofing \cite{FAS-DeSpoofing-ECCV-2018}  & 28.5  & 41.1  & \underline{33.2}  & \underline{27.8}  \bigstrut\\
		\hline
		MSR-MobileNet \cite{FAS-MSR-TIFS-2019} & 30.0  & 33.4  & \multicolumn{1}{c|}{$-$} & \multicolumn{1}{c|}{$-$} \bigstrut\\
		\hline
		MSR-ResNet \cite{FAS-MSR-TIFS-2019} \  & 36.2  & 34.7  & \multicolumn{1}{c|}{$-$} & \multicolumn{1}{c|}{$-$} \bigstrut\\
		\hline
		DRL-FAS (ours)  & \underline{28.4}  & \underline{33.2}  & \textbf{29.7}& \textbf{15.6} \bigstrut\\
		\hline
	\end{tabular}%
\end{table*}

\begin{table}[htbp]
	\centering
	\caption{The inter-database experiments where the models are trained with data of ROSE-YOUTU database and tested on the CASIA and IDIAP databases. The performance is evaluated in terms of HTER (\%). ``*'' means the results are with the outlier removal proposed in \cite{FAS-UnsupervisedDA-TIFS-2018}. On the left of $\rightarrow$ is the database used for training and the right for testing. The \textbf{best results} are highlighted in bold. In the experiments, we set $p=8$ and $T=8$.} 
	\begin{tabular}{|c|p{0.25\columnwidth}<\centering|p{0.25\columnwidth}<\centering|}
		\hline
		\multicolumn{1}{|c|}{Method} & ROSE-YOUTU$\rightarrow$CASIA & ROSE-YOUTU$\rightarrow$IDIAP \bigstrut\\
		\hline
		AlexNET without DA \cite{FAS-UnsupervisedDA-TIFS-2018}  & 32.6  & 43.6  \bigstrut\\
		\hline
		AlexNET with KMM \cite{FAS-UnsupervisedDA-TIFS-2018}  & 31.6  & 43.6  \bigstrut\\
		\hline
		AlexNET with SA \cite{FAS-UnsupervisedDA-TIFS-2018}  & 35.0  & 38.5  \bigstrut\\
		\hline
		AlexNET with KSA \cite{FAS-UnsupervisedDA-TIFS-2018}  & 33.9  & 42.0  \bigstrut\\
		\hline
		AlexNET with SA* \cite{FAS-UnsupervisedDA-TIFS-2018}  & 30.7  & 36.2  \bigstrut\\
		\hline
		AlexNET with KSA* \cite{FAS-UnsupervisedDA-TIFS-2018}  & 30.1  & 38.8  \bigstrut\\
		\hline
		DeSpoofing \cite{FAS-DeSpoofing-ECCV-2018}  & 37.2  & 38.5  \bigstrut\\
		\hline
		DRL-FAS (ours)  & \textbf{8.1}   & \textbf{20.0}  \bigstrut\\
		\hline
	\end{tabular}%
	\label{tab-rose-cross}%
\end{table}%

\par \textbf{Results on the SiW database} 
Table \ref{tab-exp-siw} shows the ACER results of our proposed framework and state-of-the-art methods on the SiW database. The Auxiliary method \cite{FAS-Auxiliary-CVPR-2018}, with extra depth map and rPPG signals, attains 1.0\%, 0.57\%, and 8.31\% ACER for Protocol 1, 2, and 3, respectively. Besides, the ``STASN+'' method \cite{FAS-CVPR2019-STASN} collects extra data outside the database for data augmentation and achieves 0.3\%, 0.15\% and 5.58\% correspondingly.
However, without extra data and auxiliary signals but only binary labels for supervision, our method can achieve the best results for the three protocols (0.00\%, 0.00\%, and 4.51\%, respectively). 
In addition, for experiments of Protocol 2 and 3, our method can manage to get the smallest standard deviation, showing better stability.
Moreover, the ACER results of all the listed methods for Protocol 3 are much higher than the results for Protocol 1 and 2.
This indicates the setting of unseen presentation attack types is more challenging than unseen faces poses and expressions as well as unseen attack mediums.

\par \textbf{Results on the OULU-NPU database}
\textcolor{black}{Table~\ref{tab-exp-oulu} compares our proposed method with state-of-the-art ones. In Protocol 1,  our method achieves 4.7\% ACER, better than the ``MSR-ResNet" method (5.9\%) \cite{FAS-MSR-TIFS-2019}, which is also based on ResNet18 \cite{CV-He-2016-ResNet}. In Protocol 2, our method achieves the best ACER (1.9\%). In Protocol 3, the Auxiliary method achieves the lowest 2.9\% ACER, while our method achieves a very close ACER of 3.0\%. In Protocol 4, our method achieves the second-best ACER of 7.2\%. Overall, in terms of ACER, the DeSpoofing is better than the Auxiliary in Protocol 1 and 4, while the Auxiliary is better than the DeSpoofing in Protocol 2 and 3. This comparison shows that there may not be a method that is always optimal for all scenarios. Nevertheless, our method shows its effectiveness by achieving the best or the second-best ACER in Protocol 2, 3, and 4. Furthermore, according to Table~\ref{tab-oulu-ext}, using a proper setting of face cropping for the framework input can lead to better performance. Also, one can always further improve the framework with advanced neural networks and auxiliary information.
}

\subsubsection{Cross-database experiments}
We also conduct cross-database experiments to evaluate the generalization capability of our method to different data domains. For conciseness, ``A $\rightarrow$B'' denotes an experiment where we run the training with the database ``A'' and run testing with the database ``B''.
\par Table \ref{tab-exp-cross-casia-msu-idiap} provides the cross-database experimental results among the CASIA, IDIAP, and MSU databases. 
In the experiments of CASIA$\rightarrow$IDIAP and IDIAP$\rightarrow$CASIA, the Auxiliary method \cite{FAS-Auxiliary-CVPR-2018} achieves the best results (27.6\% and 28.4\% HTER respectively).
On the other hand, our framework achieves second-best HTER (28.4\% and 33.2\%). Among methods without auxiliary information, such as \cite{FAS-MSR-TIFS-2019}, our performance is the best. 
Also, in both the experiments of IDIAP$\rightarrow$MSU and MSU$\rightarrow$IDIAP, we implement the DeSpoofing method \cite{FAS-DeSpoofing-ECCV-2018}, and it outperforms the other baseline methods by achieving 33.2\% and 27.8\% HTER respectively.
However, our method can significantly surpass it with much lower HTER results (29.7\% and 15.6\%), which is the best.

Table~\ref{tab-rose-cross} provides the experimental results of ROSE-YOUTU$\rightarrow$CASIA and ROSE-YOUTU$\rightarrow$IDIAP.
In the experiment of ROSE-YOUTU$\rightarrow$CASIA, our method can achieve the best 8.1\% HTER. Both the ROSE-YOUTU and CASIA database include spoofing attacks that have bezels and paper boundaries observed. Hence, the proposed framework can capture such discriminative artifacts to achieve good performance.
Besides, as for attack samples in the IDIAP database, there are few paper boundaries and bezels observed in the samples.
However, in the experiment of ROSE-YOUTU$\rightarrow$IDIAP, the proposed framework is still effective and achieves the best HTER (20.0\%), at least 16\% HTER significantly lower than the others. In summary, in the cross-database experiments, our proposed framework still shows effectiveness when the spoofing artifacts and backgrounds are from different data domains.

\begin{figure}
	\centering
	\subfigure[] {
		\label{fig-cam-a}
		\includegraphics[width=0.2\columnwidth]{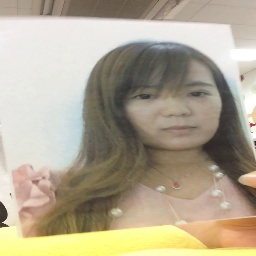}
	}
	\subfigure[] {
		\label{fig-cam-b}
		\includegraphics[width=0.2\columnwidth]{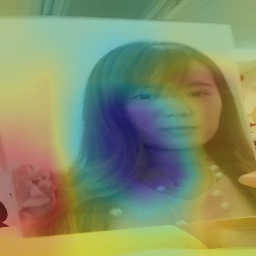}
	}
	\subfigure[ ] {
		\label{fig-cam-c}
		\includegraphics[width=0.2\columnwidth]{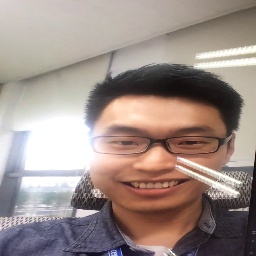}
	}
	\subfigure[] {
		\label{fig-cam-d}
		\includegraphics[width=0.2\columnwidth]{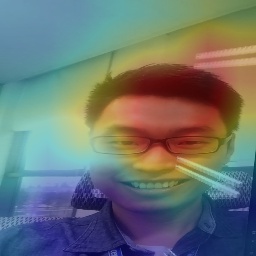}
	}
	
	\subfigure[] {
		\label{fig-cam-e}
		\includegraphics[width=0.2\columnwidth]{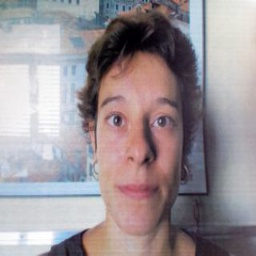}
	}
	\subfigure[] {
		\label{fig-cam-f}
		\includegraphics[width=0.2\columnwidth]{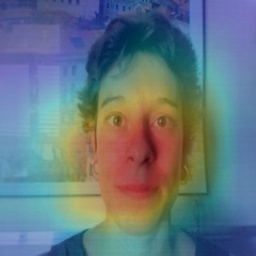}
	}
	\subfigure[] {
		\label{fig-cam-g}
		\includegraphics[width=0.2\columnwidth]{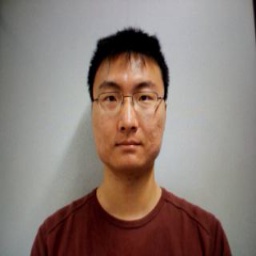}
	}
	\subfigure[] {
		\label{fig-cam-h}
		\includegraphics[width=0.2\columnwidth]{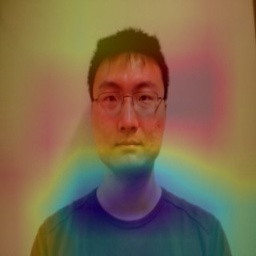}
	}
	\caption{The CAM heatmaps generated with global features. (a) is a print paper attack from the ROSE-YOUTU database, and its paper boundary is exposed on the left. (c) is a replay video attack from the ROSE-YOUTU database, and it has reflection patterns on the right. (e) and (g) are replay attack examples from the IDIAP REPLAY-ATTACK database. There are no discriminative artifacts. (b), (d), (f), and (h) are the CAM heatmaps of (a), (c), (e), and (g) respectively, where red/blue regions mean high/low activation (best viewed in color).}
	\label{fig-cam}
\end{figure}

\begin{figure*}
	\centering
	\includegraphics[width=0.95\linewidth]{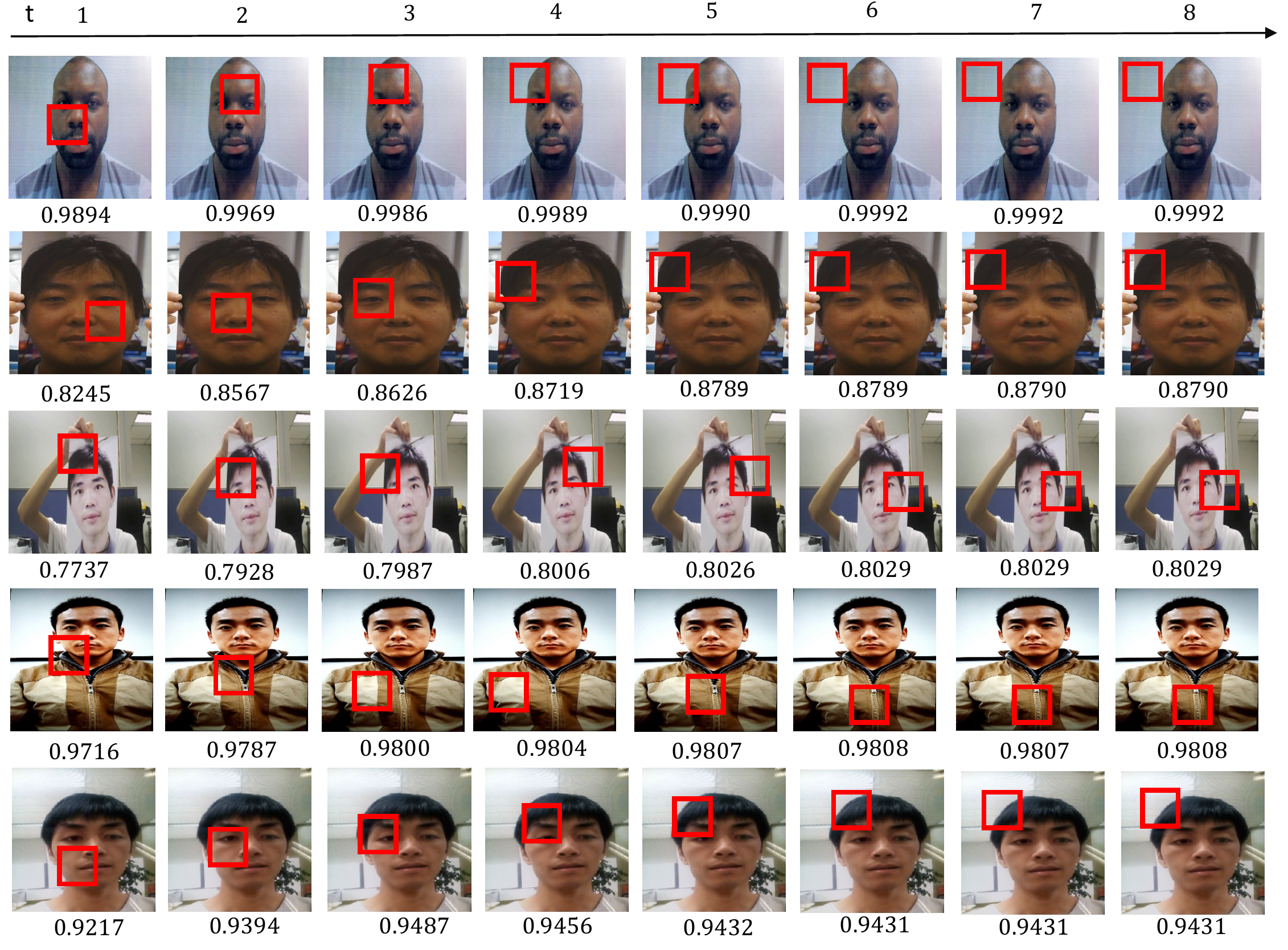}
	\caption{Visualization of each step with the confidence score. The spoofing examples are correctly classified in the intra-database experiments. Each row represents an example, and each column shows the predicted location at a step. The index of an observation step is denoted by $t$. Under each image is the confidence score $c_t$. The first row is a replay attack example from the IDIAP REPLAY-ATTACK database. The second and third rows are printed paper attack example from the CASIA database. The fourth row is a replay video attack example from the OULU-NPU database. The fifth row is a replay video attack example from the ROSE-YOUTU database.}\label{fig-vis}
\end{figure*}

\begin{figure}
	\centering
	\subfigure[CASIA] {
		\label{fig-mis-a}
		\includegraphics[width=0.25\columnwidth]{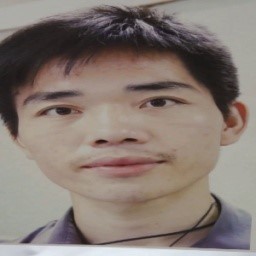}
	}
	\subfigure[ROSE-YOUTU] {
		\label{fig-mis-b}
		\includegraphics[width=0.25\columnwidth]{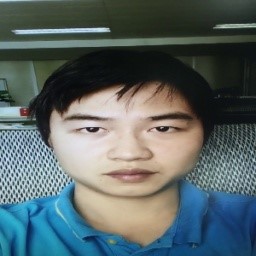}
	}
	\subfigure[OULU-NPU] {
		\label{fig-mis-c}
		\includegraphics[width=0.25\columnwidth]{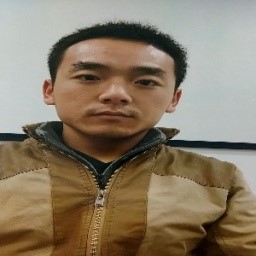}
	}
	\caption{Misclassification of spoofing examples. (a), (b), and (c) were misclassified as genuine faces in the intra-database experiments on the CAISA, ROSE-YOUTU, and OULU-NPU databases, respectively.}
	\label{fig-mis}
\end{figure}

\subsection{Visualization}

\textcolor{black}{In this subsection, we conduct further analysis based on visualization. To show what types of information the global features are likely to capture for anti-spoofing, we propose to apply the Class Activation Mapping (CAM) \cite{cam-zbl} to visualize activation heatmaps of global features. The CAM heatmaps are shown in Fig.~\ref{fig-cam}, where red/blue areas mean high/low activation. Fig.~\ref{fig-cam-a} is a paper attack where the paper boundaries can be seen on the left and the right. Its CAM heatmap, Fig.~\ref{fig-cam-b}, shows that the boundaries of both sides give high activation (red).  Fig.~\ref{fig-cam-c} is a replay video attack with reflections appearing on the right, and we see from Fig.~\ref{fig-cam-d} that the reflection areas on the top right are red. Besides, we also explore the situation when the above artifacts are absent. Fig.~\ref{fig-cam-e} and Fig.~\ref{fig-cam-g} are replay video attack examples from the IDIAP database, and there are no discriminative artifacts observed. While the face area in Fig.~\ref{fig-cam-f} gives high activation, the background areas in Fig.~\ref{fig-cam-h} are also of high activation. In summary, discriminative information captured by global features may not necessarily appear on face areas, and information from backgrounds can also significantly contribute to anti-spoofing, even when bezels, reflections, etc., are not observed.
}

\textcolor{black}{Besides, we propose to investigate how the local features can help with the performance by visualizing the predicted locations. To this end, we propose to fuse the global feature with the local feature $h_t$ extracted at the step-index $t$ ($t=1, 2...8$) for classification to get the confidence score $c_t$. In this way, the visualization results are shown in Fig.~\ref{fig-vis}, where the number under each image is the confidence score $c_t$.  
As shown in Fig.~\ref{fig-vis}, the performance of confidence scores shows an increasing trend. The first row of Fig.~\ref{fig-vis} is an attack example from the IDIAP database, a printed paper replayed in a video. As $t$ increases, the predicted patches cover the printed stripes in the background, and $c_t$ is generally increasing.
The two rows below are printed paper attack examples from the CASIA database, where the boundaries of the printed paper can be treated as discriminative spoofing artifacts, and the patches also cover the paper boundaries. In these two rows, the performance of $c_t$ shows an increasing trend. The fourth row and fifth row are replay video attack examples from the OULU-NPU database and the ROSE-YOUTU database, respectively. We observe that the local patches mainly explore moiré patterns. Meanwhile, there are $c_6>c_7$ for the fourth and $c_3>c_4$ for the fifth respectively, indicating that simply increasing $t$ may not necessarily further improve the performance. Nevertheless, for $t>1$, that $c_t>c_1$ still holds, showing that the information from patches can generally improve the overall performance. 
}
 
\textcolor{black}{Furthermore, we observe that the patches generally move from the center areas toward the boundary (background) areas.  As the initial locations are sampled from a normal distribution whose symmetry center corresponds to the center of the input image, the initial patch is generally near the center areas. Thereby, the RNN extracts features from the center areas first. As spoofing features can also be found in the background areas, driven by reinforcement learning, the patches then move toward the background areas such that the RNN can extract features from these areas to improve performance.}

\textcolor{black}{We also visualize misclassified spoofing examples from the CASIA, ROSE-YOUTU, and OULU-NPU databases, which are shown in Fig.~\ref{fig-mis}. Fig.~\ref{fig-mis}(a) is a printed paper attack from the CASIA database. Although the paper boundary at the bottom can be seen, it is not obvious, and most of the other areas have no blur or distortion observed. Fig.~\ref{fig-mis}(b) is a replay video attack example from the ROSE-YOUTU database, and Fig.~\ref{fig-mis}(c) is a replay video attack example from the OULU-NPU database. These figures show few discriminative artifacts, such as reflection. Based on the observations from these examples, as there are few discriminative artifacts observed, the extracted global and local features may not be effective in differentiating the spoofing faces from genuine ones.
}

\subsection{Computational Analysis}
\textcolor{black}{In this subsection, we analyze the computational efficiency of our proposed method and the ResNet18. As we can see from Table \ref{tab-speed}, the total number of parameters of our model is about 16.50M, while the ResNet18 is 11.18M. This increased amount of parameters is reasonable as we introduce a local branch in our framework and our model can achieve better performance in the task of FAS. Despite that our model size increases by about 50\% compared with the ResNet18, our proposed method does not introduce too much computational burden. In terms of Multiply-Accumulate-Operations (MACs) \cite{IEEE-standard}, which is for measuring the total multiplication and addition operations required for calculation, our method merely has more 0.04 Giga (1.7\%) and 0.07 (3.0\%) Giga when $T=4$ and $T=8$ than the ResNet18 baseline.  Last but not least, we consider the inference efficiency by reporting Frames per Second (FPS) based on PyTorch and NVIDIA GTX 1080 Ti GPU. As we can see, while the ResNet18 achieves 150 FPS, our method can achieve 110 FPS and 70 FPS when $T=4$ and $T=8$, which is reasonable as the local branch works recurrently. Nevertheless, in practice, one can always use various neural network acceleration techniques to speed up models.}

\begin{table}[t]
	\centering
	\caption{The computational information of ResNet18 and our method.}
	\begin{tabular}{|l|c|c|c|c|}
		\hline
		Method & Params & MACs & FPS \bigstrut\\
		\hline
		ResNet18 & 11.18M & 2.38G & 150 \bigstrut\\
		\hline
		Ours (T=4) &16.50M & 2.42G & 110 \bigstrut\\
		\hline
		Ours (T=8) &16.50M & 2.45G & 70 \bigstrut\\
		\hline
	\end{tabular}%
	\label{tab-speed}%
\end{table}%

\section{Conclusion} \label{sec-conclusion}
We present a novel two-branch framework to explore spoofing clues for face anti-spoofing problem. The novelties of our work lie in two folds, 1) we propose to leverage CNN and RNN to extract both global and local information for the FAS task based on a single frame; 2) we propose a novel optimization strategy based on deep reinforcement learning, which is the first attempt in the FAS problem. We conduct extensive experiments on six different databases to evaluate our proposed framework. The experimental results on both intra- and cross-domain indicate that our proposed framework can generally achieve state-of-the-art performance compared with various state-of-the-art baselines, which demonstrate the effectiveness of our method.


\ifCLASSOPTIONcaptionsoff
  \newpage
\fi


\bibliographystyle{ieeetr}
\bibliography{main.bib}

\end{document}